\documentclass[lettersize,journal]{IEEEtran}
\usepackage{amsmath,amsfonts,amsthm}
\usepackage{algorithmic}
\usepackage{algorithm}
\usepackage{array}
\usepackage[caption=false,font=normalsize,labelfont=sf,textfont=sf]{subfig}
\usepackage{textcomp}
\usepackage{stfloats}
\usepackage{url}
\usepackage{verbatim}
\usepackage{graphicx}
\usepackage{cite}
\usepackage{booktabs}
\usepackage{multirow}
\usepackage{makecell}
\usepackage{color}
\usepackage{CJKutf8}
\usepackage{pifont}
\usepackage{colortbl}
\newcolumntype{H}{>{\setbox0=\hbox\bgroup}c<{\egroup}@{}}
\usepackage{bm}
\usepackage{bbm}
\usepackage{pythonhighlight}
\usepackage{mathrsfs}
\usepackage{mathrsfs}
\usepackage{MnSymbol}
\usepackage{dashrule}
\usepackage{tikz}
\usepackage{mathtools}
\usepackage{amsthm}

\usepackage{adjustbox}
\usepackage{pifont}

\usepackage[utf8]{inputenc} 
\usepackage[T1]{fontenc}    
\usepackage{hyperref}       
\usepackage{url}            
\usepackage{booktabs}       
\usepackage[table]{xcolor} 
\usepackage{bm} 
\usepackage{amsfonts}       
\usepackage{nicefrac}       
\usepackage{microtype}      
\usepackage{xcolor}         
\usepackage{siunitx} 
\usepackage{subcaption}
\usepackage{wrapfig}
\usepackage{multirow}
\usepackage{multicol}
\usepackage{booktabs}       
\usepackage{graphicx}
\usepackage{amsmath}
\usepackage{array}
\usepackage[table,xcdraw]{xcolor}
\usepackage{color}
\usepackage{algorithm}
\usepackage{algorithmic}
\usepackage{tabularx}
\usepackage{hyperref}     
\usepackage{booktabs}   
\usepackage{soul}          
\usepackage{xcolor}        

\begin{document}

\title{PixelFlowCast: Latent-Free Precipitation Nowcasting via \\Pixel Mean Flows}

\author{Yufeng~Zhu$^{\dagger}$, Chunlei~Shi$^{\dagger}$, Yongchao~Feng, and Dan~Niu$^{\ddagger}$

\thanks{This work was supported by the  Heavy Rainfall Research Foundation of China (No. BYKJ2025M14), China Meteorological Administration Xiong\textrm{'}an Atmospheric Boundary Layer Key Laboratory (No. 2025LABL-B12), and by the National Natural Science Foundation of China (62374031, 62331009), and by NSFC-Jiangsu Province (BK20240173). (\textit{$^{\dagger}$Yufeng Zhu and $^{\dagger}$Chunlei Shi have contributed equally to this work. $^{\ddagger}$Corresponding author: Dan Niu.})}
\thanks{Yufeng Zhu, Chunlei Shi and Dan Niu are with the Department of Automation, Southeast University, Nanjing 210096, China (e-mail: 220252049@seu.edu.cn, 230238514@seu.edu.cn, danniu1@163.com).}
\thanks{Yongchao Feng is with the State Key Laboratory of Virtual Reality Technology and Systems, Beihang University, Beijing 100191, China.}}

\markboth{Journal of \LaTeX\ Class Files,~Vol.~14, No.~8, August~2021}%
{Shell \MakeLowercase{\textit{et al.}}: A Sample Article Using IEEEtran.cls for IEEE Journals}


\maketitle

\begin{abstract}
Precipitation nowcasting aims to forecast short-term radar echo sequences for extreme weather warning, where both prediction fidelity and inference efficiency are critical for real-world deployment. However, diffusion-based models, despite their strong generative capability, suffer from slow inference due to multi-step sampling trajectories, limiting their practical usability. Conditional Flow Matching (CFM) improves efficiency via straightened trajectories, but relies on latent space compression, which inevitably discards high-frequency physical details and degrades fine-grained prediction quality. To address these limitations, we propose PixelFlowCast, a two-stage probabilistic forecasting framework that achieves both high-efficiency and high-fidelity prediction without latent compression. Specifically, in the first stage, a deterministic model first produces coarse forecasts to capture global evolution trends. In the subsequent stage, the proposed KANCondNet extracts deep spatiotemporal evolution features to provide accurate conditional guidance. Based on this, a latent-free, few-step Pixel Mean Flows (PMF) predictor employs an $x$-prediction mechanism to generate high-quality predictions, effectively preserving fine-grained structures while maintaining fast inference. Experiments on the publicly available SEVIR dataset demonstrate that PixelFlowCast outperforms existing mainstream methods in both prediction accuracy and inference efficiency, particularly for long sequence forecasting, highlighting its strong potential for real-world operational deployment.
\end{abstract}


\section{Introduction}
\label{sec:Introduction}
Precipitation nowcasting aims to predict the short-term evolution of rainfall over a target region for the next few hours 
by analyzing meteorological data~\cite{NIPS2015_07563a3f, NEURIPS2020_fa78a161, catão2025precipitationnowcastingsatellitedata}. 
The accuracy and efficiency of these forecasts are of paramount importance for issuing timely and reliable extreme weather warnings~\cite{NKUNZIMANA2020105069, 10.3389/feart.2022.846113}. 
Existing deep learning models for nowcasting can be broadly categorized into deterministic and probabilistic generative approaches~\cite{LI2026130775}. 
Deterministic models~\cite{NIPS2017_a6db4ed0, Gao_2022_CVPR, NEURIPS2022_a2affd71, wang2022predrnnrecurrentneuralnetwork} directly predict future radar echoes based on historical data. 
However, as they typically optimize for Mean Squared Error (MSE), long-term forecasts tend to produce overly smooth and blurry images, 
lacking microscopic details and authenticity.


Probabilistic generative models have emerged as a promising solution to this smoothing problem. 
Among these, diffusion models~\cite{NEURIPS2020_4c5bcfec, Rombach_2022_CVPR} show great promise due to their high fidelity and training stability. 
PreDiff~\cite{NEURIPS2023_f82ba6a6} combines latent diffusion with spatiotemporal representations to improve perceptual quality, 
while DiffCast~\cite{Yu_2024_CVPR} and CasCast~\cite{gong2024cascastskillfulhighresolutionprecipitation} adopt two-stage architectures 
that effectively balance macroscopic trends with microscopic details. 
Despite these visual improvements, diffusion models fundamentally rely on complex curved denoising trajectories and multi-step sampling iterations. 
This inherent multi-step process causes high inference latency, 
severely restricting their practical deployment in real-time meteorological warning systems.

\begin{figure}[t]
    \centering
    \includegraphics[width=1.0\linewidth, trim=0pt 0pt 0pt 0pt, clip]{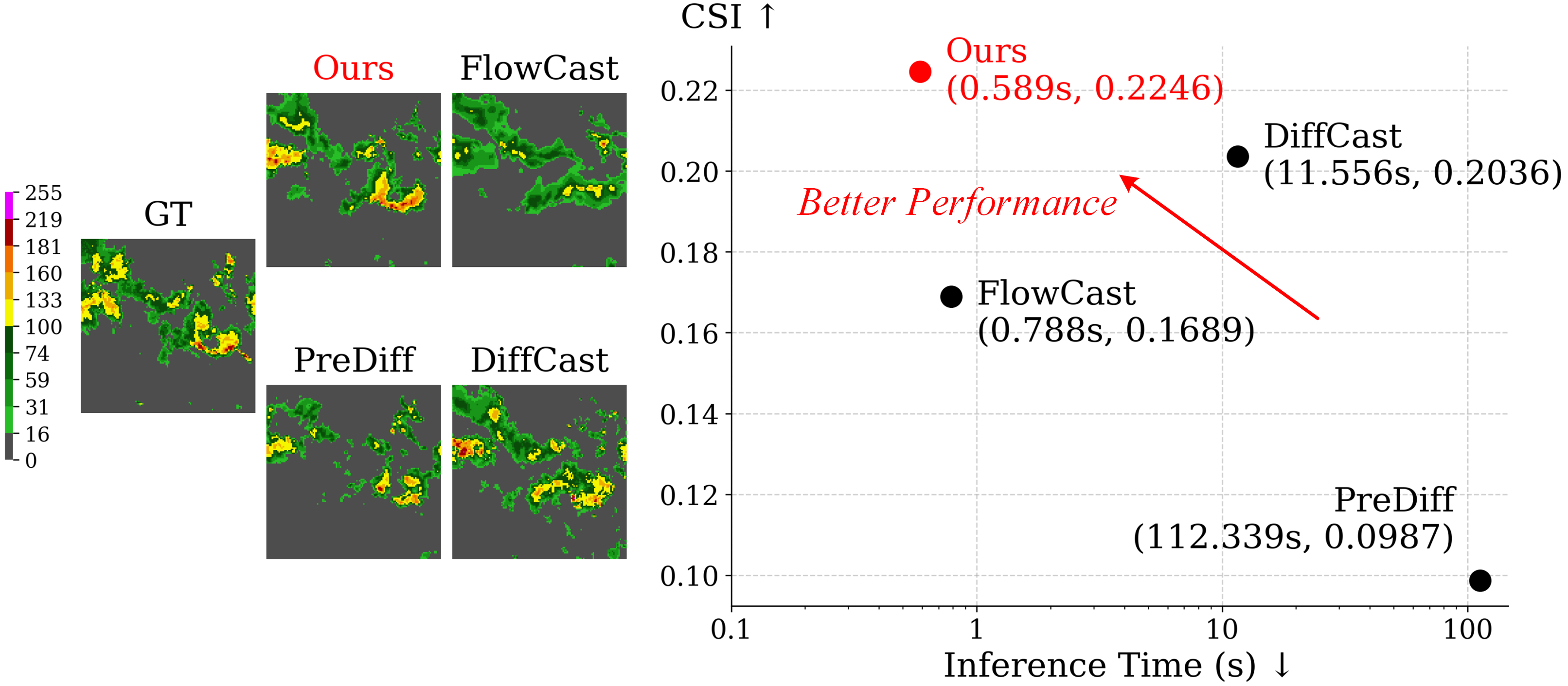}
    \caption{Performance versus efficiency of precipitation nowcasting models. 
    Existing generative models suffer from multi-step sampling latency and detail loss due to latent compression. 
    Our PixelFlowCast achieves superior fidelity and efficiency through latent-free, few-step prediction.}
    \label{fig:motivation}
\vspace{-3mm}
\end{figure}

Driven by the critical need for faster inference, recent approaches like FlowCast~\cite{ribeiro2026flowcastadvancingprecipitationnowcasting} 
utilize Conditional Flow Matching (CFM)~\cite{lipman2023flowmatchinggenerativemodeling, tong2024improvinggeneralizingflowbasedgenerative, liu2022flowstraightfastlearning} 
to learn a straightened evolutionary trajectory, significantly reducing the sampling steps required by traditional diffusion models. 
While these CFM-based models successfully improve efficiency, they typically rely on latent space compression to perform dimensionality reduction.
This spatial compression inevitably strips away high-frequency physical dynamics, such as the localized intensity of extreme heavy precipitation. 
Furthermore, simply discarding the latent space to preserve these microscopic details is unfeasible for standard CFM, 
because directly regressing complex velocity fields in a high-dimensional pixel space often leads to severe optimization difficulties~\cite{lu2026onesteplatentfreeimagegeneration}.
Consequently, as illustrated in Figure~\ref{fig:motivation}, existing generative paradigms face a critical dilemma: 
forcing a strict compromise between real-time inference efficiency and high-fidelity detail preservation.

To overcome the optimization hurdles of operating in high-dimensional spaces without lossy latent compression, 
the Pixel Mean Flows (PMF) mechanism~\cite{lu2026onesteplatentfreeimagegeneration, geng2025meanflowsonestepgenerative} 
offers a promising perspective. 
Originally proposed for general image generation, PMF formulates the network as an $x$-prediction model 
based on the generalized manifold hypothesis~\cite{10.1145/1390156.1390294}. 
This bypasses direct velocity field regression, inherently facilitating latent-free, accelerated sampling directly in the pixel space. 
However, directly transferring this accelerated sampling to complex meteorological data risks introducing a slight spatial smoothing effect, 
potentially degrading the prediction of high-threshold extreme weather events.

Fortunately, Kolmogorov-Arnold Networks (KANs)~\cite{liu2025kankolmogorovarnoldnetworks, bodner2025convolutionalkolmogorovarnoldnetworks, drokin2024kolmogorovarnoldconvolutionsdesignprinciples, 10.1145/3743128} 
have recently emerged as an ideal complement to counteract this detail degradation. 
By replacing fixed activation functions with learnable splines, KANs excel at approximating complex nonlinear mappings. 
While recent studies~\cite{11052773, CHENG2025134134} have validated their ability to capture local microscopic dynamics 
and enhance high-threshold forecasting in deterministic models, 
their potential to guide and enhance probabilistic generative frameworks has yet to be unleashed.

Inspired by the paradigm shift from direct velocity regression to $x$-prediction for efficient latent-free generation, 
and the deployment of learnable spline functions over fixed activations for complex nonlinear mapping, 
we propose a novel two-stage probabilistic generative forecasting framework, PixelFlowCast. 
Instead of modeling the full evolution directly, PixelFlowCast decomposes the forecasting task. 
A deterministic model is first utilized to establish a macroscopic spatiotemporal baseline. 
To generate the missing microscopic physical details, we design a conditional encoding network, KANCondNet, 
which leverages the nonlinear expressiveness of learnable splines to extract deep spatiotemporal evolution features. 
Conditioned on this precise guidance, we propose the PMF predictor to execute an accelerated, latent-free sampling process, 
directly generating high-frequency residuals via $x$-prediction. 
By superimposing these generative details onto the smooth coarse baseline, 
our model yields the final high-fidelity forecast. As illustrated in Figure~\ref{fig:motivation}, 
this unified design successfully breaks the bottleneck of existing generative paradigms, 
achieving real-time operational efficiency without sacrificing critical meteorological details. Our main contributions are summarized as:
\begin{itemize}
    \item We propose PixelFlowCast, a novel two-stage generative framework that operates directly in the pixel space to achieve latent-free, high-fidelity, and fast precipitation nowcasting.
    \item We propose KANCondNet to extract deep spatiotemporal evolution features, and the latent-free PMF predictor to generate high-frequency residuals via $x$-prediction under the guidance of KANCondNet.
    \item Experiments on the SEVIR dataset show that PixelFlowCast outperforms current baselines in generation quality and inference speed.
\end{itemize}

\section{Related Work}
\subsection{Deterministic Predictive Models }

Deterministic models laid the foundation for data-driven precipitation nowcasting. 
Early studies like ConvLSTM~\cite{shi2015convolutionallstmnetworkmachine} integrate convolutions into Recurrent Neural Networks (RNNs). 
TrajGRU~\cite{NIPS2017_a6db4ed0} adapts to precipitation system deformations by dynamically learning location-variant connections. 
Subsequently, Earthformer~\cite{NEURIPS2022_a2affd71} introduces a space-time Transformer architecture to capture global dependencies, 
while SimVP~\cite{Gao_2022_CVPR} proposes a purely convolutional end-to-end architecture to improve computational efficiency. 
Although these models effectively capture macroscopic motion, the inherent smoothing effect of MSE loss functions often 
leads to blurry predictions and the loss of microscopic physical details.

\subsection{Probabilistic Generative Models}

Probabilistic generative models are introduced to address the over-smoothing defect. 
Generative Adversarial Networks (GANs) ~\cite{8761462, ravuri2021skilful, zhang2023skilful, meo2024extremeprecipitationnowcastingusing} can restore precipitation textures but face training instability. 
In contrast, diffusion models guarantee stable training by optimizing an evidence lower bound (ELBO). 
Ldcast~\cite{leinonen2023latentdiffusionmodelsgenerative} and PreDiff~\cite{NEURIPS2023_f82ba6a6} 
reduce computational dimensions using latent spaces, while DiffCast~\cite{Yu_2024_CVPR} and CasCast~\cite{gong2024cascastskillfulhighresolutionprecipitation} 
adopt a two-stage architecture of deterministic coarse prediction and generative refinement. 
However, constrained by the reverse Markov process, diffusion models require multi-step denoising iterations, 
causing high computational latency that conflicts with real-time nowcasting requirements.

Consequently, recent flow-based generative models like FlowCast~\cite{ribeiro2026flowcastadvancingprecipitationnowcasting} 
successfully reduce inference steps by employing CFM to learn straightened mapping trajectories.
To alleviate the high computational cost of modeling high-dimensional spatiotemporal data, 
FlowCast depends on Variational Autoencoders (VAEs)~\cite{kingma2022autoencodingvariationalbayes, oord2018neuraldiscreterepresentationlearning} for latent space compression. 
Unfortunately, this lossy dimensionality reduction inevitably causes the smoothing and loss of microscopic physical details. 
Furthermore, standard flow matching cannot simply discard the latent space to preserve these details, as directly regressing continuous velocity fields 
in high-dimensional pixel space causes severe optimization difficulties~\cite{lu2026onesteplatentfreeimagegeneration}.

\subsection{Kolmogorov-Arnold Networks}

KANs replace traditional Multi-Layer Perceptrons (MLPs) by deploying learnable spline functions on network edges, 
significantly improving the approximation accuracy of complex nonlinear mappings~\cite{liu2025kankolmogorovarnoldnetworks}. In precipitation nowcasting, 
EvoKAN~\cite{CHENG2025134134} introduces the ConvKAN module to radar echo extrapolation to alleviate spatial smoothing. 
SwinKAN~\cite{11052773} utilizes a dual-branch architecture to integrate the large-scale global capturing of Swin Transformers 
with the local microscopic focusing of ConvKAN. 
While previous studies primarily apply KANs to the feature extraction layers of deterministic models for nonlinear fitting, 
our study introduces KANCondNet into a generative framework. 
By extracting deep spatiotemporal evolution features and providing precise conditional guidance, 
it effectively improves the prediction performance for high-threshold heavy precipitation.

\begin{figure*}[t]
    \centering
    \includegraphics[width=1\textwidth, trim=1pt 0pt 0pt 0pt, clip]{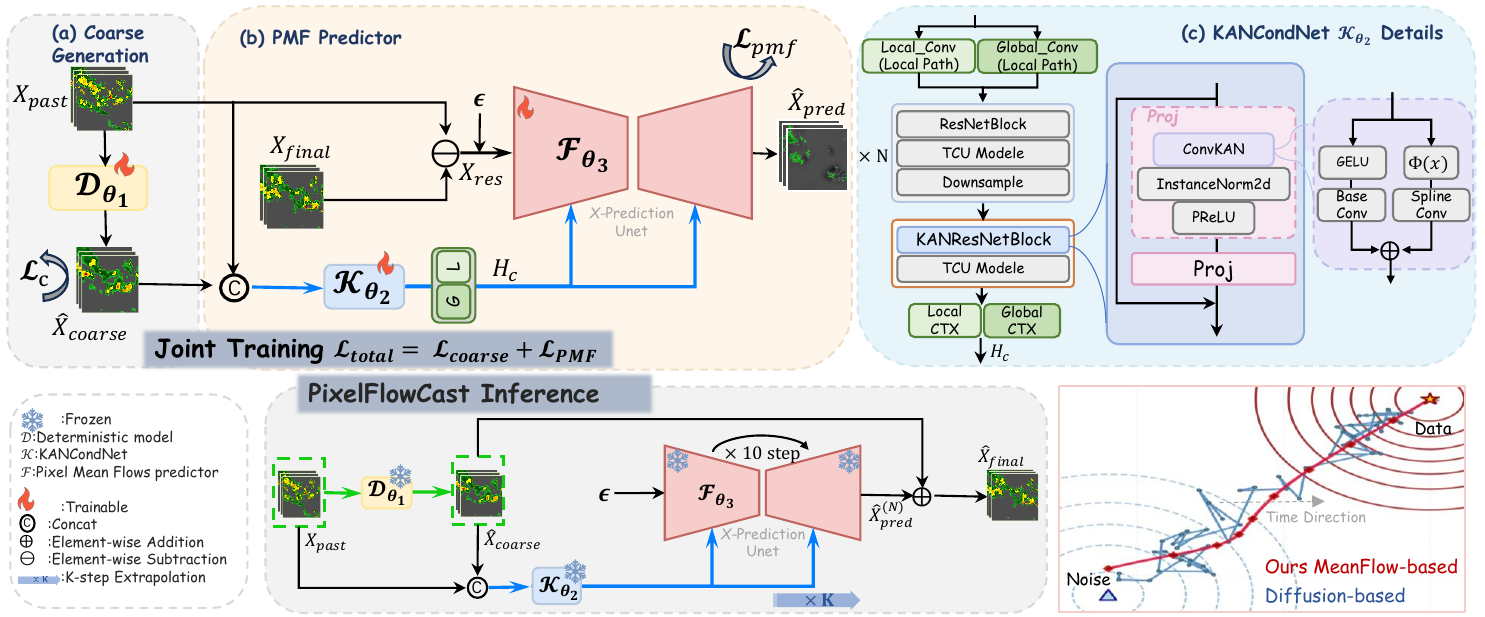}
    
    \caption{The overview of our PixelFlowCast framework and its core component KANCondNet. 
    In Stage 1, a deterministic model generates a macroscopic coarse prediction $\hat{X}_{coarse}$ from historical radar echoes $X_{past}$. 
    In Stage 2, the proposed KANCondNet (detailed in the right panel) extracts precise spatiotemporal conditions from the combined inputs 
    to guide the PMF predictor, which directly generates high-frequency residuals in the pixel space within few steps. 
    The final high-fidelity prediction $\hat{X}_{final}$ is obtained by superimposing the residual $\hat{X}_{res}$ (i.e., $\hat{X}_{pred}^{(N)}$) onto the coarse baseline $\hat{X}_{coarse}$. 
    }
    \label{fig:overall_framework}
\end{figure*}

\section{Method}

\subsection{Task Formulation}
\label{sec:task_formulation}
Precipitation nowcasting via radar echo extrapolation can be defined as a spatiotemporal sequence prediction problem, 
namely using the past $T_{in}$ radar echo frames $X_{past} = \{X_1, X_2, \dots, X_{T_{in}}\}$ 
to predict the future $T_{out}$ frames $X_{future} = \{X_{T_{in}+1}, X_{T_{in}+2}, \dots, X_{T_{in}+T_{out}}\}$, 
where $X_t \in \mathbb{R}^{H \times W \times C}$. 
Formally, the proposed PixelFlowCast framework is defined as a mapping function $\mathcal{P}_{\theta}$ parameterized by network weights $\theta$. 
The overall precipitation nowcasting problem can be compactly formulated 
as generating the predicted future sequence $\hat{X}_{final}$ based on the historical observations:
\begin{equation}
\hat{X}_{final} = \mathcal{P}_{\theta}(X_{past}).
\label{eq:task_formulation}
\end{equation}
The primary objective in learning $\mathcal{P}_{\theta}$ is to capture 
both macroscopic evolution and high-frequency microscopic details, 
while maintaining high inference efficiency without relying on lossy latent compressions.

\subsection{Overall Framework}
\label{sec:Overall Framework}

As illustrated in Figure~\ref{fig:overall_framework}, we propose PixelFlowCast, 
a novel two-stage probabilistic generative forecasting framework. 
Inspired by the paradigm shift to $x$-prediction for efficient latent-free generation, 
our core motivation is to achieve accelerated sampling directly in the original pixel space 
without the lossy latent compression that strips away high-frequency physical dynamics. 
Rather than generating highly dynamic meteorological systems from scratch, 
PixelFlowCast decomposes the forecasting task into a deterministic coarse prediction stage and a latent-free, few-step generative refinement stage.

In the first stage, a deterministic model receives the real radar echo sequence $X_{past}$ of the past $T_{in}$ frames 
to establish a macroscopic spatiotemporal baseline. 
It captures the macroscopic motion trend, outputting a coarse prediction $\hat{X}_{coarse}$ for the target prediction window. 
The missing microscopic physical details can therefore be formulated as the target residual $X_{res}$ 
between the actual future observation $X_{future}$ and this coarse baseline $\hat{X}_{coarse}$.

Addressing this high-frequency residual and counteracting the detail degradation typically faced in high-dimensional generation requires robust condition encoding. 
Consequently, $X_{past}$ and $\hat{X}_{coarse}$ are concatenated along the channel dimension to serve as the input for KANCondNet. 
By deploying learnable spline functions over fixed activations for complex nonlinear mapping, 
KANCondNet effectively extracts deep spatiotemporal evolution features. 
Utilizing these extracted features as precise meteorological conditions, 
the PMF predictor bypasses direct velocity field regression via an $x$-prediction model. 
This inherently facilitates a latent-free, accelerated sampling mechanism directly in the pixel space 
to produce the predicted residual $\hat{X}_{res}$. 
The final high-fidelity prediction is then obtained by summing the outputs of the two stages:
\begin{equation}
    \hat{X}_{final} = \hat{X}_{coarse} + \hat{X}_{res}.
\end{equation}

It is worth noting that the formulation above represents a single temporal prediction window of length $T_{window}$. 
In our implementation, we set $T_{window} = T_{in}$ to maintain a consistent temporal receptive field. 
To extend this capability to long-term sequence forecasting of length $T_{out}$, 
the entire framework is encapsulated within an autoregressive generation wrapper consisting of $K = T_{out} / T_{window}$ steps. 
By updating the input sequence $X_{past}$ in a rolling manner using the high-fidelity predictions from the current window, 
the model continuously extrapolates the spatiotemporal evolution into the extended future. 
For clarity, Sections~\ref{sec:KANCondNet} and \ref{sec:Pixel Mean Flows predictor} will focus exclusively on the inner workings of a single prediction window, 
and the specific algorithmic details of this autoregressive mechanism will be fully elaborated in Section~\ref{sec:Training and Inference}.

\subsection{KANCondNet}
\label{sec:KANCondNet}
To bridge the gap between the overly smooth macroscopic baseline and the highly dynamic physical reality, 
the PMF predictor requires precise spatiotemporal guidance to generate high-frequency residuals. 
Therefore, we design a conditional encoding network, KANCondNet, aimed at extracting deep evolutionary features 
from both past contexts and future trends.

Meteorological systems, particularly extreme precipitation events, exhibit highly complex and nonlinear spatiotemporal dynamics. 
Traditional convolutions, constrained by fixed activation functions, lack the adaptive representational capacity to resolve such chaotic, 
high-frequency variations. 
Motivated by this, KANCondNet leverages the KAN architecture. 
It enhances representational capacity by placing learnable spline functions on network edges rather than using fixed node activations. 
Based on the Kolmogorov-Arnold representation theorem~\cite{liu2025kankolmogorovarnoldnetworks}, 
a smooth multivariate function $f: [0,1]^n \to \mathbb{R}$ is decomposed into a finite composition of univariate continuous functions:
\begin{equation}
    f(\mathbf{x}) = f(x_1, \dots, x_n) = \sum_{q=1}^{2n+1} \Phi_{q} \left( \sum_{p=1}^{n} \phi_{q,p}(x_p) \right),
\end{equation}
where $\mathbf{x} \in [0, 1]^n$ is the $n$-dimensional input, 
and $x_p$ denotes its $p$-th component. The terms $\phi_{q,p}: [0,1] \to \mathbb{R}$ and $\Phi_{q}: \mathbb{R} \to \mathbb{R}$ 
denote the inner and outer univariate continuous functions, respectively.

To adapt this theoretical decomposition for high-dimensional spatiotemporal modeling, 
the architecture is instantiated as the Convolutional KAN (ConvKAN). 
It parameterizes the connections within the local receptive field using B-splines rather than standard scalar weights, 
which enhances the representational capacity of spatial features while maintaining translational invariance.
Following the design principles of the existing ConvKAN~\cite{drokin2024kolmogorovarnoldconvolutionsdesignprinciples}, 
the forward propagation of the single-layer features in our ConvKAN is instantiated as a dual-path aggregation mechanism that 
combines base mapping and spline mapping in parallel:
\begin{equation}
    H_{out} = \mathbf{W}_{\mathrm{base}} \ast \sigma(H_{in}) + \mathbf{W}_{\mathrm{spline}} \ast \mathrm{Spline}(H_{in}),
\end{equation}
where $\ast$ denotes the convolution operation, $\sigma$ is the base activation function, 
and $\text{Spline}(\cdot)$ represents the learnable B-spline basis functions aggregated over the feature grid.

Specifically, the combined input of $X_{past}$ and $\hat{X}_{coarse}$ is initially mapped to a high-dimensional space 
via a standard Conv2d layer and then passes through cascaded downsampling KANResnetBlocks 
and a Temporal Cues Unit (TCU)~\cite{li2024scrd} to extract spatiotemporal features. 
Ultimately, this hierarchical extraction process outputs a set of multi-scale condition tensors $H_{c} = \{h_1, h_2, \dots, h_L\}$. 

KAN architectures exhibit exceptional parameter efficiency and representational capacity in capturing meteorological fluid dynamics~\cite{CHENG2025134134, 11052773}. 
Because of this strong expressiveness, blindly deploying KANResnetBlocks throughout the network is not only unnecessary, 
but also risks overfitting to the chaotic, high-frequency noise inherent in precipitation systems. 
Therefore, we strategically confine the deployment of KANResnetBlocks to the deepest layer of KANCondNet, 
while allowing the shallower layers to revert to standard ResNetBlocks. 
This configuration ensures that KANCondNet focuses its spline-based learning precisely on highly compressed spatiotemporal features, 
thereby striking an balance between nonlinear representation capacity and robust generalization.

Furthermore, to ensure long-term forecasting robustness within the overarching autoregressive framework, 
KANCondNet is designed to extract conditions dynamically. 
Rather than treating the combined inputs as static representations across time, 
the module flexibly leverages the error-free local context from historical observations 
and the broad global guidance from the coarse forecast to construct the precise multi-scale condition $H_c$ for the current prediction window:
\begin{equation}
    H_c = \text{KANCondNet}(X_{past}, \hat{X}_{coarse}).
\end{equation}

\begin{figure*}[t]
    \begin{minipage}[t]{0.49\textwidth}
        \begin{algorithm}[H]
        \caption{Training of PixelFlowCast}
        \label{alg:training}
        \begin{algorithmic}[1]
        \REQUIRE Past radar frames $X_{past}$, future ground truth $X_{future}$, total autoregressive steps $K$, parameters $\theta$
        \WHILE{not converged}
            \STATE Sample $(X_{past}, X_{future})$ from dataset
            \STATE $\hat{X}_{coarse} \leftarrow \text{DeterministicModel}(X_{past})$
            \STATE Compute $\mathcal{L}_{coarse} = \text{MSE}(\hat{X}_{coarse}, X_{future})$
            \STATE Sample a random autoregressive step index $k \sim \mathcal{U}\{1, K\}$
            \STATE $X_{res}^{(k)} \leftarrow X_{future}^{(k)} - \hat{X}_{coarse}^{(k)}$ \quad \algorithmiccomment{Target residual}
            \STATE $H_c \leftarrow \text{KANCondNet}(X_{past}^{(k)}, \hat{X}_{coarse}^{(k)})$ 
            \STATE Sample $t \sim \mathcal{U}(0, 1)$, $r \sim \mathcal{U}(0, t)$, and $\epsilon \sim \mathcal{N}(0, \mathbf{I})$
            \STATE $Z_t \leftarrow (1-t)X_{res}^{(k)} + t\epsilon$ \quad \algorithmiccomment{Construct forward path}
            \STATE $\hat{X}_{pred} \leftarrow F_\theta(Z_t, r, t, H_c)$ \quad \algorithmiccomment{x-prediction}
            \STATE $u_\theta \leftarrow (Z_t - \hat{X}_{pred}) / t$ \quad \algorithmiccomment{Convert to average velocity}
            \STATE $v_\theta \leftarrow u_\theta + (t-r) \frac{d u_\theta}{d t}$ \quad \algorithmiccomment{Derive instantaneous velocity}
            \STATE Compute $\mathcal{L}_{PMF} = \| v_\theta - (\epsilon - X_{res}^{(k)}) \|_2^2$ 
            \STATE $\mathcal{L}_{total} \leftarrow \mathcal{L}_{coarse} + \mathcal{L}_{PMF}$
            \STATE Update parameters $\theta$ using gradient descent on $\nabla_\theta \mathcal{L}_{total}$
        \ENDWHILE
        \end{algorithmic}
        \end{algorithm}
    \end{minipage}
    \hfill
    \begin{minipage}[t]{0.49\textwidth}
        \begin{algorithm}[H]
        \caption{Inference of PixelFlowCast}
        \label{alg:inference}
        \begin{algorithmic}[1]
        \REQUIRE Input radar frames $X_{past}$, total autoregressive steps $K$, sampling steps $N$
        \STATE $\hat{X}_{coarse} \leftarrow \text{DeterministicModel}(X_{past})$
        \FOR{$k = 1$ to $K$}
            \STATE $H_c \leftarrow \text{KANCondNet}(X_{past}^{(k)}, \hat{X}_{coarse}^{(k)})$
            \STATE Initialize $Z_{curr} \sim \mathcal{N}(0, \mathbf{I})$
            \STATE Set step size $\Delta t \leftarrow 1 / N$
            \FOR{$i = 0$ to $N-1$}
                \STATE $t \leftarrow 1 - i \cdot \Delta t$ \quad \algorithmiccomment{Current time}
                \STATE $r \leftarrow t - \Delta t$ \quad \algorithmiccomment{Target time for average velocity}
                \STATE $\hat{X}_{pred} \leftarrow F_\theta(Z_{curr}, r, t, H_c)$ \quad \algorithmiccomment{x-prediction}
                \STATE $u \leftarrow (Z_{curr} - \hat{X}_{pred}) / t$ \quad \algorithmiccomment{Average velocity}
                \STATE $Z_{curr} \leftarrow Z_{curr} - u \cdot \Delta t$ \quad \algorithmiccomment{Euler step}
            \ENDFOR
            \STATE $\hat{X}_{res}^{(k)} \leftarrow \hat{X}_{pred}$ \quad \algorithmiccomment{Final residual generated by model $F_{\theta}$}
            \STATE Update $X_{past}^{(k+1)}$ via sliding window with $(\hat{X}_{coarse}^{(k)} + \hat{X}_{res}^{(k)})$ 
        \ENDFOR
        \STATE $\hat{X}_{res} \leftarrow \text{Concat}(\hat{X}_{res}^{(1)}, \dots, \hat{X}_{res}^{(K)})$ \quad \algorithmiccomment{Combine all residuals}
        \STATE $\hat{X}_{final} \leftarrow \hat{X}_{coarse} + \hat{X}_{res}$ \quad \algorithmiccomment{Final refinement}
        \RETURN $\hat{X}_{final}$
        \end{algorithmic}
        \end{algorithm}
    \end{minipage}
\end{figure*}

\subsection{Pixel Mean Flows predictor}
\label{sec:Pixel Mean Flows predictor}
While the deterministic first stage successfully captures macroscopic evolution trends, 
it inherently produces overly smooth baselines lacking realistic meteorological details. 
Therefore, the second stage is designed to explicitly generate these critical high-frequency residuals. 
To achieve fast inference directly in the high-dimensional pixel space without relying on lossy latent compression, 
our framework employs the PMF predictor.
Unlike traditional flow matching that tightly couples the prediction and optimization spaces by directly regressing a continuous velocity field, 
the PMF predictor explicitly decouples them to enable stable, latent-free generation directly in the pixel space.
During training, the core optimization process consists of three steps:

\textbf{Forward Process.}
Given pure noise $\epsilon \sim \mathcal{N}(0, \mathbf{I})$ and the target ground-truth residual $X_{res}$, 
the forward process constructs a straightforward linear interpolation path. 
For any current time step $t \in [0, 1]$ (where $t=1$ represents pure noise and $t=0$ represents the clean target), 
the intermediate noisy state $Z_t$ is defined as:
\begin{equation}
    Z_t = (1-t)X_{res} + t\epsilon.
\end{equation}

\textbf{$\boldsymbol{X}$-Prediction \& Average Velocity.} 
Directly regressing abstract velocity fields for chaotic radar echoes is difficult. 
To overcome this, the PMF predictor formulates $F_\theta$ as an $x$-prediction model parameterized by an auxiliary time step $r \sim \mathcal{U}(0, t)$, 
which represents an earlier state to define the average velocity interval and capture the evolutionary trajectory. 
Guided by the condition $H_c$ from KANCondNet, the network directly generates the target residual via $x$-prediction as:
\begin{equation}
    \hat{X}_{pred} = F_\theta(Z_t, r, t, H_c). 
    \label{eq:x_pred} 
\end{equation}
While this explicit pixel-space prediction $\hat{X}_{pred}$ provides a stable anchor, 
integrating it into the continuous flow matching framework requires a mathematical conversion into an average velocity field $u_\theta$:
\begin{equation}
    u_\theta(Z_t, r, t, H_c) = \frac{Z_t - \hat{X}_{pred}}{t}.
    \label{eq:u_theta} 
\end{equation}

\textbf{Conditional PMF Objective.}
While the prediction operates in the data space, the optimization must be performed in the instantaneous velocity space 
to maintain stable flow matching gradients. 
Using the MeanFlow identity, the instantaneous velocity $v_\theta$ is derived from the average velocity $u_\theta$ via the Jacobian-vector product (JVP):
\begin{equation}
    v_\theta(Z_t, r, t, H_c) = u_\theta(Z_t, r, t, H_c) + (t-r) \frac{d u_\theta(Z_t, r, t, H_c)}{d t}.
    \label{eq:jvp} 
\end{equation}
The loss objective is then designed to minimize the discrepancy between this derived $v_\theta$ and the ground-truth instantaneous velocity $\epsilon - X_{res}$:
\begin{equation}
    \mathcal{L}_{PMF} = \mathbb{E}_{t, r, \epsilon, X} \left[ \| v_\theta(Z_t, r, t, H_c) - (\epsilon - X_{res}) \|_2^2 \right]. 
    \label{eq:loss_pmf} 
\end{equation}
This dual-space formulation, which combines prediction in the original pixel space with optimization in the velocity space, 
ensures that the generated output tightly converges to the target $X_{res}$ without requiring any latent compression.

\textbf{Accelerated Inference via $\boldsymbol{X}$-Prediction.}
During inference, the $x$-prediction paradigm enables an efficient few-step trajectory approximation, 
significantly reducing the sampling steps required by traditional diffusion models. 
Initialized with pure noise $Z_{1}=\epsilon$ at $t=1$, the model performs a few-step Euler sampling process. 
At each step, the network directly predicts the target residual $\hat{X}_{pred}$, 
which is then used to compute the average velocity and update the intermediate state $Z_{curr}$. 
Crucially, integrating chaotic velocity fields across discrete steps often accumulates numerical errors in high-dimensional pixel space. 
To circumvent this instability, rather than returning the accumulated state $Z_{0}$, 
our mechanism directly extracts the bounded $x$-prediction from the final $N$-th step 
(denoted as $\hat{X}_{pred}^{(N)}$) as the generated residual $\hat{X}_{res}$:
\begin{equation}
    \hat{X}_{res} = \hat{X}_{pred}^{(N)} = \text{Sampler}(F_{\theta}, \epsilon, H_c, N),
\end{equation}
where $N$ represents the minimal number of sampling steps.

\subsection{Training and Inference}
\label{sec:Training and Inference}

\textbf{Joint Training Objective.}
During training, the two stages of PixelFlowCast are optimized simultaneously. The total loss function is defined as:
\begin{equation} 
    \mathcal{L}_{total} = \mathcal{L}_{coarse} + \mathcal{L}_{PMF}, 
\end{equation}
where $\mathcal{L}_{coarse}$ is the MSE loss for the first stage, 
and $\mathcal{L}_{PMF}$ is the loss defined in Section~\ref{sec:Pixel Mean Flows predictor} for the second stage.

\textbf{Autoregressive Sequence Generation.}
To forecast long-term sequences of length $T_{out}$, the framework operates autoregressively over $K = T_{out} / T_{window}$ steps. 
We use the superscript $(k)$ (where $k \in \{1, \dots, K\}$) to index variables within the $k$-th prediction window. 
During this continuous progression, the input sequence $X_{past}^{(k)}$ (which strictly maintains a fixed length of $T_{in}$) 
is updated step-by-step through a sliding window mechanism. 
Specifically, in each autoregressive step, the $T_{window}$ oldest frames are shifted 
out of the conditioning input, and the $T_{window}$ newly generated predictions are appended 
to form the temporal context for the subsequent step.

Detailed procedures for training and inference are summarized in Algorithm~\ref{alg:training} and Algorithm~\ref{alg:inference}, respectively.

\begin{table*}[t]
    \centering
    \caption{Quantitative comparison of PixelFlowCast with baseline models on the SEVIR dataset. ``Overall Average'' represents metrics computed over the full 3-hour forecast (36 frames), while ``Last 1 Hour'' focuses exclusively on the 2-3 hour prediction phase. Best results are highlighted in \textbf{bold} and second-best are \underline{underlined}.}
    \label{tab:sota_comparison_sevir}
    \begin{tabular*}{\textwidth}{@{\extracolsep{\fill}} l cccccc cc @{}}
        \toprule
        & \multicolumn{6}{c}{Overall Average (All Thresholds, 3 Hours)} & \multicolumn{2}{c}{Last 1 Hour} \\
        \cmidrule(lr){2-7} \cmidrule(lr){8-9}
        Model & CSI $\uparrow$ & CSI-pool4 $\uparrow$ & CSI-pool16 $\uparrow$ & HSS $\uparrow$ & LPIPS $\downarrow$ & SSIM $\uparrow$ & CSI $\uparrow$ & HSS $\uparrow$ \\
        \midrule
        U-Net~\cite{NEURIPS2020_fa78a161}       & 0.1877 & 0.2028 & 0.2173 & 0.2267 & 0.4207 & \underline{0.5929} & \underline{0.1435} & 0.1747 \\
        Earthformer~\cite{NEURIPS2022_a2affd71} & 0.2006 & 0.2193 & 0.2384 & 0.2456 & 0.4434 & 0.5773 & 0.1366 & 0.1666 \\
        SimVP~\cite{Gao_2022_CVPR}       & \underline{0.2073} & \underline{0.2277} & \underline{0.2405} & 0.2518 & 0.4348 & 0.5667 & 0.1362 & 0.1640 \\
        \midrule
        PreDiff~\cite{NEURIPS2023_f82ba6a6}     & 0.0987 & 0.1057 & 0.0974 & 0.1310 & 0.3666 & 0.5774 & 0.0246 & 0.0305 \\
        FlowCast~\cite{ribeiro2026flowcastadvancingprecipitationnowcasting}    & 0.1689 & 0.1760 & 0.1855 & 0.2087 & 0.3927 & 0.4435 & 0.1210 & 0.1489 \\
        DiffCast~\cite{Yu_2024_CVPR}    & 0.2036 & 0.2191 & 0.2299 & \underline{0.2620} & \underline{0.3592} & 0.5315 & 0.1399 & \underline{0.1791} \\
        \midrule
        \textbf{PixelFlowCast (Ours)} & \textbf{0.2246} & \textbf{0.2424} & \textbf{0.2476} & \textbf{0.2867} & \textbf{0.3241} & \textbf{0.6106} & \textbf{0.1500} & \textbf{0.1907} \\
        \bottomrule
    \end{tabular*}
\end{table*}

\begin{table*}[t]
    \centering
    \caption{Detailed performance comparison of PixelFlowCast and generative baselines across different VIL thresholds on the SEVIR dataset. All metrics are averaged over the full 3-hour forecast horizon (36 frames). Inference time is measured as seconds per sample. Best results are highlighted in \textbf{bold} and second-best are \underline{underlined}.}
    \label{tab:threshold_comparison}
    \begin{tabular*}{\textwidth}{@{\extracolsep{\fill}} ll cccccc c @{}}
        \toprule
        \multirow{2}{*}{\textbf{Metric}} & \multirow{2}{*}{\textbf{Model}} & \multicolumn{6}{c}{\textbf{VIL Thresholds}} & \multirow{2}{*}{\textbf{Time (s) $\downarrow$}} \\
        \cmidrule(lr){3-8}
        & & \textbf{16} & \textbf{74} & \textbf{133} & \textbf{160} & \textbf{181} & \textbf{219} & \\
        \midrule
        \multirow{4}{*}{CSI $\uparrow$} 
        & PreDiff~\cite{NEURIPS2023_f82ba6a6} & 0.2874 & 0.1635 & 0.0657 & 0.0350 & 0.0268 & 0.0138 & 112.339 \\
        & FlowCast~\cite{ribeiro2026flowcastadvancingprecipitationnowcasting} & \underline{0.5832} & 0.3552 & 0.0388 & 0.0185 & 0.0129 & 0.0048 & \underline{0.788} \\ 
        & DiffCast~\cite{Yu_2024_CVPR} & 0.5635 & \underline{0.4163} & \underline{0.1320} & \underline{0.0544} & \underline{0.0381} & \underline{0.0172} & 11.556 \\
        & \textbf{PixelFlowCast (Ours)} & \textbf{0.6041} & \textbf{0.4552} & \textbf{0.1541} & \textbf{0.0636} & \textbf{0.0470} & \textbf{0.0239} & \textbf{0.589} \\
        \midrule
        \multirow{4}{*}{HSS $\uparrow$} 
        & PreDiff~\cite{NEURIPS2023_f82ba6a6} & 0.3325 & 0.2193 & 0.1036 & 0.0597 & 0.0463 & 0.0247 & 112.339 \\
        & FlowCast~\cite{ribeiro2026flowcastadvancingprecipitationnowcasting} & \underline{0.6463} & 0.4727 & 0.0671 & 0.0335 & 0.0237 & 0.0091 & \underline{0.788} \\
        & DiffCast~\cite{Yu_2024_CVPR} & 0.6295 & \underline{0.5347} & \underline{0.2131} & \underline{0.0948} & \underline{0.0677} & \underline{0.0321} & 11.556 \\
        & \textbf{PixelFlowCast (Ours)} & \textbf{0.6701} & \textbf{0.5747} & \textbf{0.2434} & \textbf{0.1079} & \textbf{0.0809} & \textbf{0.0429} & \textbf{0.589} \\
        \bottomrule
    \end{tabular*}
\end{table*}
\section{Experiments}
\label{sec:experiments}

\subsection{Experimental Setting}
\label{sec:experimental_setting}

\textbf{Dataset.} 
This study employs the Vertically Integrated Liquid (VIL) field from the widely 
used SEVIR dataset in the field of meteorological deep learning for experimental evaluation. 
This dataset captures the evolution of radar echoes within a $384 \times 384 \text{ km}$ domain, 
with a spatial resolution of 1 km and a temporal resolution of 5 minutes. Following the official standards of the SEVIR dataset, 
we use a chronological split to divide the events into independent training, validation, and test sets. 
The training set covers the period from June 2017 to December 2018, containing 24,076 sequence samples; 
the validation set spans from January 2019 to September 2019, with 5,703 sequence samples; and the test set is from October 
2019 to November 2019, totaling 1,466 sequence samples. 

\textbf{Data Preprocessing.} 
Due to the limitation of computing resources, 
during the preprocessing stage, we downsampled the spatial dimensions of all original radar frames 
to $128 \times 128$ pixels. In the construction of the time series, 
each sample is a 48-frame sequence extracted from the continuous radar observations. 
The model receives the radar observation fields from the past 1 hour (i.e., 12 consecutive frames) as the input context, 
aiming to forecast the spatiotemporal evolution of the precipitation system for the next 3 hours (i.e., 36 consecutive frames).

\textbf{Evaluation Metrics.} 
Following DiffCast~\cite{Yu_2024_CVPR}, 
we evaluate nowcasting accuracy using the Critical Success Index (CSI) and Heidke Skill Score (HSS) at standard VIL thresholds (16, 74, 133, 160, 181, 219). 
To account for minor spatial deviations, we report CSI at $4 \times 4$ and $16 \times 16$ pooling scales. 
Additionally, LPIPS and SSIM are utilized to measure the visual quality of the predictions.
Furthermore, to evaluate the real-time operational capability of our latent-free framework, 
we measure the average inference time per complete prediction sequence. Evaluations are conducted on the first 10\% 
of the SEVIR test set using a single GPU (batch size 1), following a 5-sample warm-up.


\subsection{Training and Inference Details}
\label{sec:Training and Inference Details}

\textbf{Training.} 
We employ SimVP~\cite{Gao_2022_CVPR} as the first-stage deterministic model of PixelFlowCast 
for joint training and optimize the model using the AdamW optimizer with an initial learning rate of $1 \times 10^{-4}$ 
and a cosine annealing scheduler with a 10\% linear warmup. For the optimization objective, we set the weights of the coarse prediction loss $\mathcal{L}_{coarse}$
and the PMF predictor loss $\mathcal{L}_{PMF}$ to a 1:1 ratio and train the model on 4 NVIDIA GeForce RTX 3090 (24GB) GPUs for 100 epochs. 
Within KANCondNet, we specifically deploy a single KANResnetBlock at the deepest layer.

\textbf{Inference.} 
As described in Section \ref{sec:Pixel Mean Flows predictor}, 
the proposed PixelFlowCast adopts an accelerated few-step sampling mechanism, 
specifically setting the number of integration steps to $N=10$ when generating prediction results.

Further details regarding datasets, evaluation metrics, and network configurations are provided in the Appendix.

\subsection{Comparison with State-of-the-Art}
\label{sec:Comparison with State-of-the-Art}

We evaluate the proposed PixelFlowCast against two categories of representative baseline models: 
deterministic predictive models, U-Net~\cite{NEURIPS2020_fa78a161}, Earthformer~\cite{NEURIPS2022_a2affd71}, SimVP~\cite{Gao_2022_CVPR} 
and probabilistic generative models PreDiff~\cite{NEURIPS2023_f82ba6a6}, DiffCast~\cite{Yu_2024_CVPR}, FlowCast~\cite{ribeiro2026flowcastadvancingprecipitationnowcasting}. 
For all evaluated baseline models, we utilize their official open-source implementations 
and train them to convergence on the target dataset using their default hyperparameters. 
For the inference of the generative baselines, 
FlowCast utilizes a 10-step Euler ODE solver, DiffCast employs the DDIM sampler with 250 denoising steps, 
and PreDiff adopts the standard DDPM generation paradigm involving 1000 denoising steps.

\begin{figure}[t]
    \centering
    \includegraphics[width=0.8\linewidth, trim=10 5 10 5, clip]{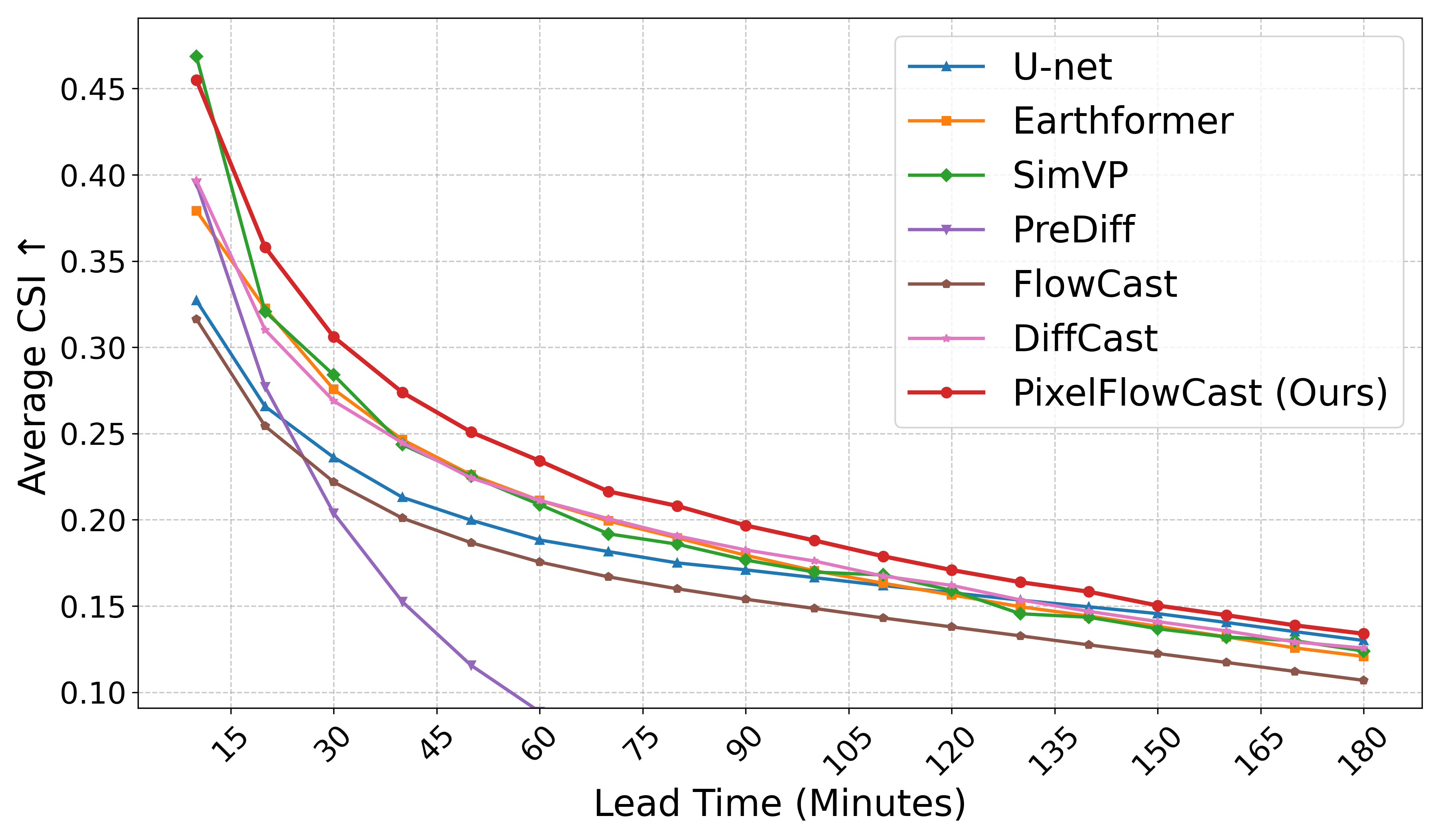}
    \vspace{-2mm} 
    \caption{Average CSI degradation over a 3-hour forecast lead time (10-minute intervals).}
    \label{fig:temporal_csi}
\end{figure}

\begin{figure}[t]
    \centering
    \includegraphics[width=0.8\linewidth, trim=10 5 10 5, clip]{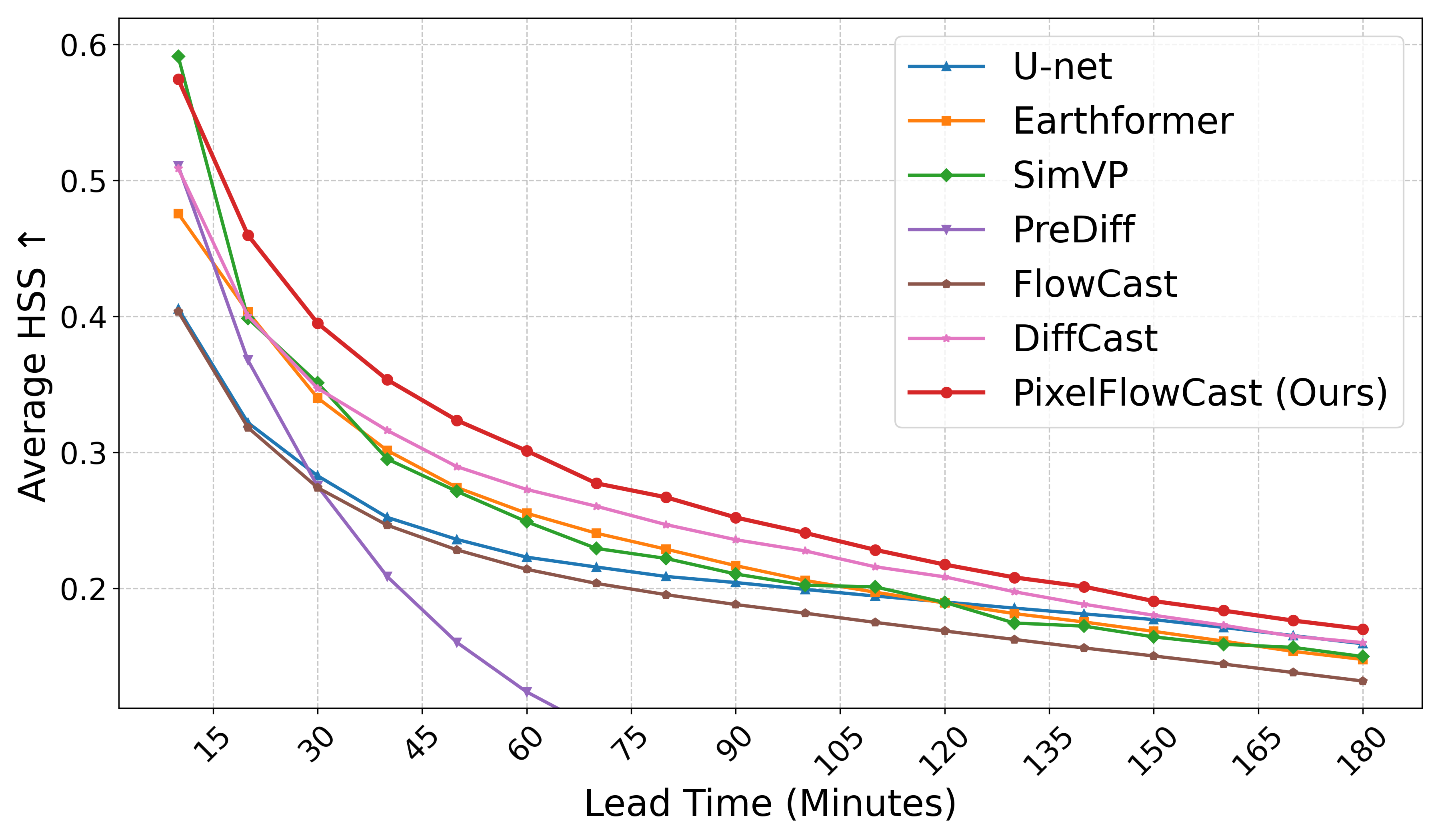}
    \vspace{-2mm}
    \caption{Average HSS degradation over a 3-hour forecast lead time (10-minute intervals).}
    \label{fig:temporal_hss}
\end{figure}


\begin{figure}[t]
    \centering
    \includegraphics[width=0.98\linewidth, trim=8pt 8pt 8pt 8pt, clip]{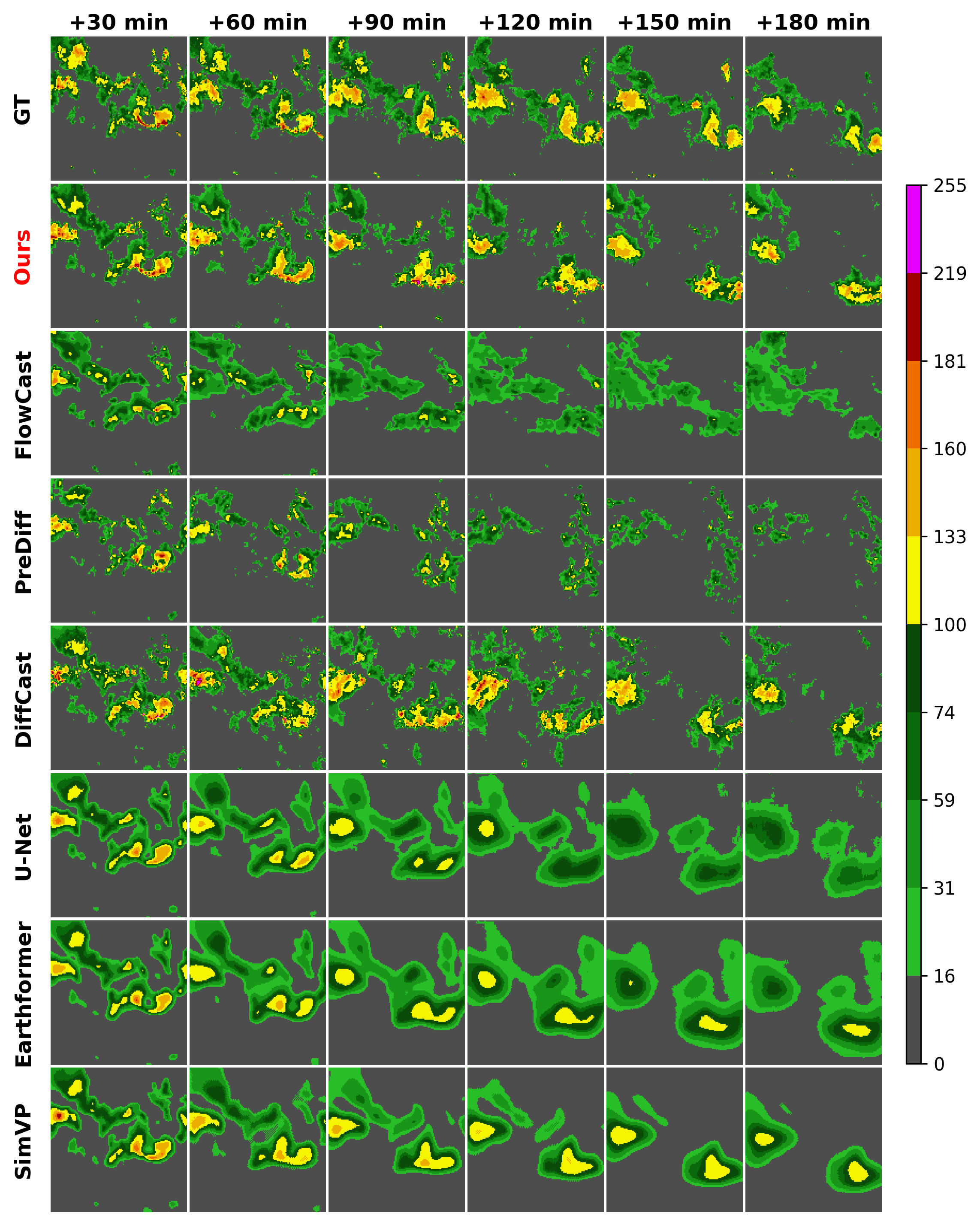}
    \caption{A set of example forecasts on the SEVIR dataset. Predictions are displayed at 30-minute intervals up to a 180-minute lead time.}
    \label{fig:qualitative_comparison}
\vspace{-5mm}
\end{figure}

\begin{figure*}[t] 
    \centering
    
    \begin{minipage}[b]{0.48\textwidth}
        \centering
        \includegraphics[trim=1.2cm 1.5cm 1.5cm 1.5cm, clip, width=0.75\linewidth]{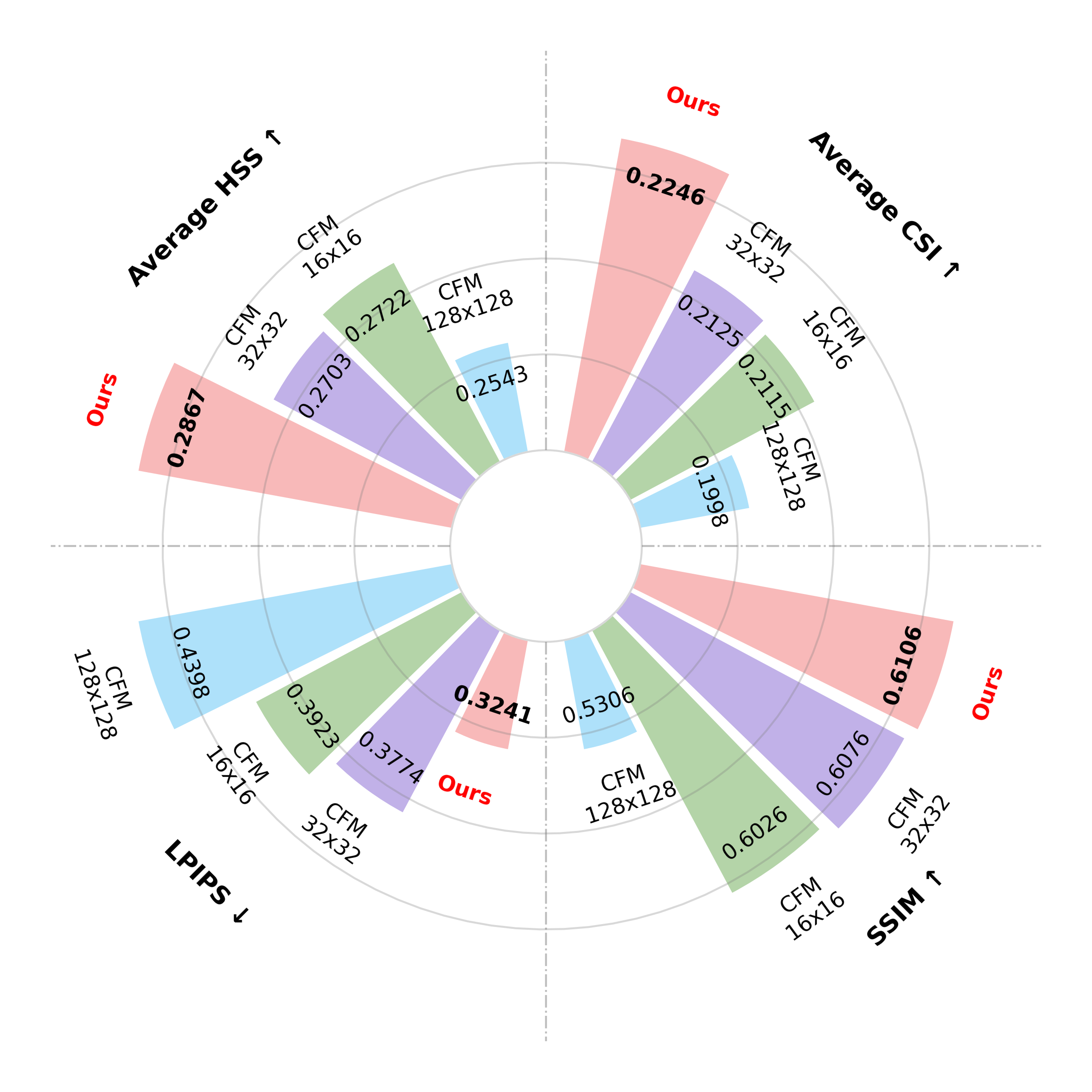}
        \caption{Ablation study on PMF generative paradigm.}
        \label{fig:ablation_pmf_vs_fm}
    \end{minipage}
    \hfill 
    \begin{minipage}[b]{0.48\textwidth}
        \centering
        \includegraphics[trim=1.1cm 1.1cm 1.1cm 1.1cm, clip, width=0.75\linewidth]{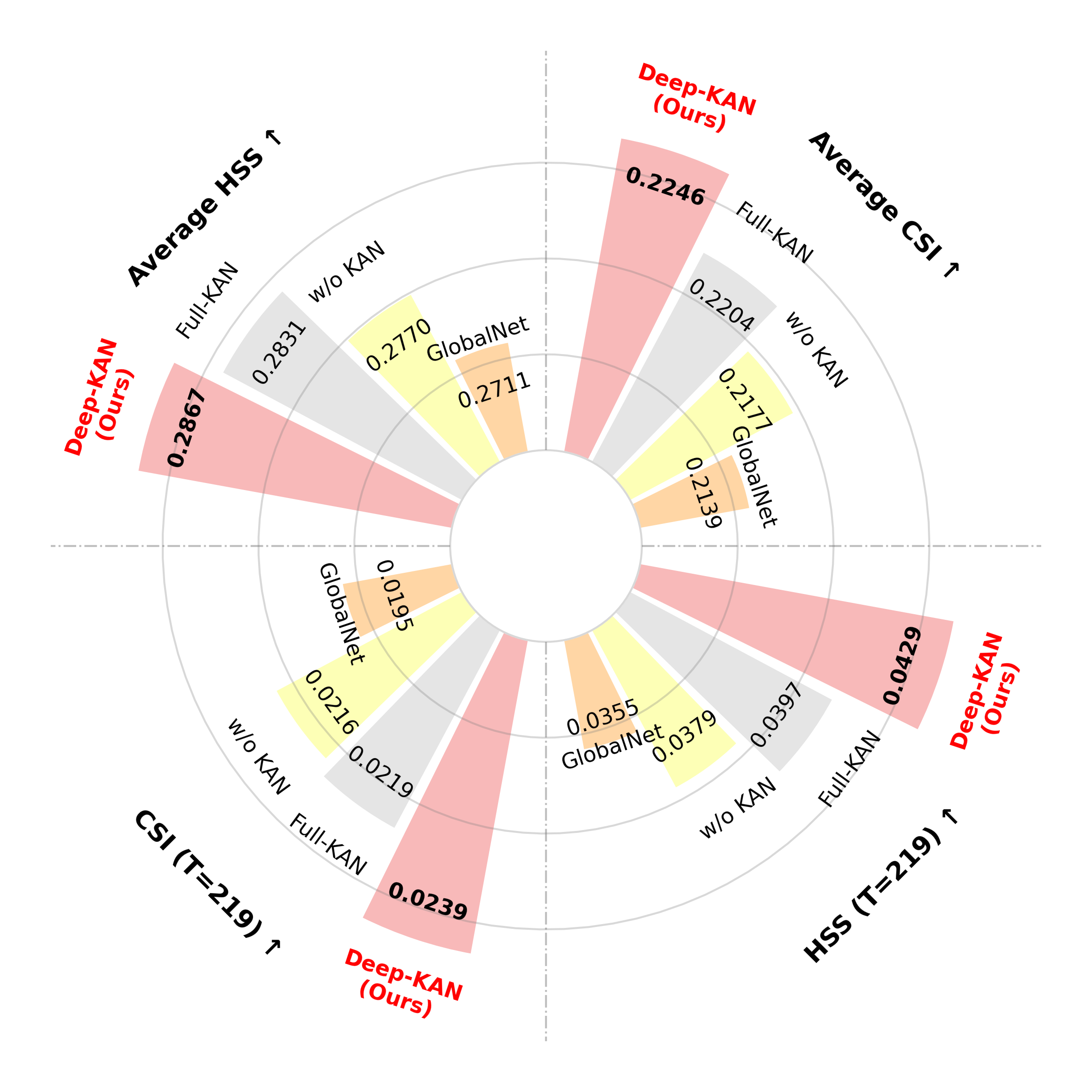}
        \caption{Ablation study of KANCondNet module.}
        \label{fig:ablation_kancondnet}
    \end{minipage}

\end{figure*}
As shown in Table~\ref{tab:sota_comparison_sevir}, PixelFlowCast achieves an optimal balance between prediction accuracy and visual quality, 
consistently outperforming both deterministic and generative baselines. Regarding macroscopic accuracy, 
it reaches an overall CSI of 0.2246 and HSS of 0.2867, yielding relative improvements of 8.3\% and 13.9\% over the best deterministic baseline (SimVP). 
This performance gap is even more pronounced when compared to purely generative models like FlowCast and PreDiff, which lack a deterministic physical foundation, 
causing their metrics to decline significantly over the 3-hour forecast horizon. Furthermore, while the two-stage DiffCast similarly incorporates deterministic constraints, 
PixelFlowCast maintains a clear performance advantage throughout the entire period, achieving the best LPIPS (0.3241) and SSIM (0.6106). 
This confirms that our approach of discarding latent compression in favor of accelerated pixel-space prediction effectively 
avoids the smoothing effects of deterministic models while preserving the structural integrity that pure generative methods often lack.

Table~\ref{tab:threshold_comparison} further details the performance of the generative models across different VIL thresholds, along with their average inference times. 
While predictive performance naturally declines for all models as the evaluation threshold increases, 
PixelFlowCast maintains a consistent performance advantage, achieving the best CSI and HSS across all six thresholds. 
Most notably, at extreme heavy precipitation thresholds (VIL $\ge$ 133), our model significantly outperforms the strong two-stage diffusion baseline, DiffCast. 
This sensitivity to extreme weather events demonstrates the effectiveness of KANCondNet. 
Its strong nonlinear representation capability accurately extracts deep spatiotemporal dynamics, 
successfully avoiding the smoothing compromises typical of latent-space compressions. 

Beyond predictive accuracy, Table~\ref{tab:threshold_comparison} also demonstrates that PixelFlowCast effectively resolves the critical fidelity-efficiency trade-off. 
Compared to the severe computational latency of traditional diffusion models (PreDiff at 112.34s, DiffCast at 11.56s) and even the flow-based FlowCast (0.788s), 
PixelFlowCast requires only 0.589s per sample. 
By executing an accelerated few-step sampling mechanism (i.e., 10 steps) directly in the original pixel space, 
our framework delivers highly competitive, detail-rich extreme precipitation forecasts while fully satisfying strict real-time operational requirements.

Furthermore, in terms of long-term forecasting, 
PixelFlowCast achieves the highest CSI (0.1500) and HSS (0.1907) during the challenging ``Last 1 Hour'' phase (Table~\ref{tab:sota_comparison_sevir}).
This sustained superiority in mitigating error accumulation is further illustrated by the temporal degradation curves 
in Figures~\ref{fig:temporal_csi} and \ref{fig:temporal_hss}, which show our model consistently suppressing performance decay 
across the entire 180-minute forecasting horizon. 
Visually, these quantitative achievements are corroborated by Figure~\ref{fig:qualitative_comparison}: 
while traditional baselines suffer from severe blurring and structural decay at later lead times, 
PixelFlowCast preserves sharp precipitation boundaries and accurate spatial distributions even at the 180-minute mark. 
Together, these results confirm the effectiveness of our framework in capturing and retaining complex meteorological dynamics over extended periods.

\subsection{Ablation Studies}
\label{sec:Ablation Studies}


\textbf{Generative Paradigm Comparison.} 
To verify the effectiveness of making predictions directly in the original pixel space without latent compression, 
we replace the core PMF algorithm with standard CFM without changing the overall model framework. 
As illustrated in the radar chart (Figure~\ref{fig:ablation_pmf_vs_fm}), 
if standard CFM is applied directly in the original pixel space ($128 \times 128$) without using VAE (i.e., CFM $128 \times 128$), 
the prediction performance is suboptimal. Its overall CSI drops to 0.1998, and the image structure metric SSIM decreases noticeably to 0.5306. 
When VAE is introduced for spatial compression, the macroscopic performance recovers. 
The Latent CFM models operating in $16 \times 16$ and $32 \times 32$ latent spaces achieve comparable results, 
with their overall CSI increasing to 0.2115 and 0.2125, respectively. 
However, they still consistently lag behind the PMF approach.
This phenomenon corroborates the argument raised in Section~\ref{sec:Introduction} that standard CFM models suffer from severe optimization difficulties 
when directly regressing velocity fields in high-dimensional pixel spaces, forcing them to rely on VAE compression that inevitably smooths physical details.
In contrast, by completely discarding latent compression and generating high-frequency residuals via $x$-prediction, 
the PMF predictor successfully preserves microscopic physical details in the original pixel space, 
delivering the highest CSI (0.2246), SSIM (0.6106), and an optimal LPIPS (0.3241).

\textbf{Effectiveness Evaluation of KANCondNet.} 
The evaluation of KANCondNet begins with a comparison against GlobalNet (the conditional encoder in DiffCast~\cite{Yu_2024_CVPR}) to verify its representational capacity. 
As shown in the radar chart (Figure~\ref{fig:ablation_kancondnet}), KANCondNet consistently outperforms GlobalNet, 
yielding relative improvements of 22.6\% in CSI and 20.8\% in HSS at the extreme precipitation threshold ($T=219$).
This confirms that KAN-based nonlinear modeling provides more effective features for capturing the localized microscopic dynamics.

To further investigate the optimal application of KANResNetBlocks, 
we conduct an ablation study on the number of KANResNetBlocks within KANCondNet. 
We compare three deployment strategies: w/o KAN (using standard ResNetBlocks exclusively), 
Full-KAN (deploying KANResNetBlocks across all spatial stages), and our Deep-KAN (confining KANResNetBlock strictly to the deepest layer).
While incorporating KANResNetBlocks at any stage surpasses the w/o KAN baseline, 
Full-KAN fails to yield further gains and even shows slight degradation in CSI. 
This supports our hypothesis proposed in Section~\ref{sec:KANCondNet} that 
pervasive KAN deployment may lead to nonlinear overfitting on chaotic, high-frequency noise.
Consequently, our Deep-KAN achieves the best balance between nonlinear expressiveness and robust generalization,
demonstrating that confining learnable splines to deep, abstract feature levels is the most effective architectural choice.
\section{Conclusion}
\label{sec:conclusion}

In this paper, we propose PixelFlowCast, an efficient two-stage generative framework for precipitation nowcasting. 
Precipitation nowcasting fundamentally remains challenging due to the 
highly chaotic nature of atmospheric systems and the strict demands for timely processing.
To address this, our PixelFlowCast employs KANCondNet to extract deep spatiotemporal evolution features, 
providing precise guidance for a latent-free PMF predictor. 
By operating directly in the pixel space via an $x$-prediction formulation, 
the proposed approach bypasses latent compression and multi-step generation, 
inherently balancing accelerated sampling with fine-grained fidelity for high-threshold extreme weather events.
Experiments on the SEVIR dataset demonstrate that PixelFlowCast outperforms 
existing mainstream baselines in both prediction accuracy and inference efficiency for long sequence forecasting.
In future work, we aim to incorporate explicit atmospheric physical laws 
into the few-step generation process to further enhance the physical consistency of the predictions.

\bibliographystyle{IEEEtran}
\bibliography{egbib}

\clearpage
\appendix
\label{appendix}

\renewcommand{\thefigure}{S-\arabic{figure}}
\renewcommand{\thetable}{S-\arabic{table}}
\renewcommand{\theequation}{S-\arabic{equation}}

This document provides supplementary material for the main manuscript. The contents are organized as follows:
\textbf{Appendix~\ref{supp:Extended Formulation of PixelFlowCast}}
elaborates on the extended formulation of the PixelFlowCast framework, including geometric interpretations and the multi-step sampling strategy.
\textbf{Appendix~\ref{supp:Experimental Settings and Implementation Details}} 
provides comprehensive experimental settings and implementation details, covering dataset preprocessing, training configurations, and specific network architectures.
\textbf{Appendix~\ref{sec:additional_ablation}} 
presents additional ablation studies focusing on inference efficiency, sampling steps, and the noise-free extraction strategy.
\textbf{Appendix~\ref{sec:Limitations and Future Work}} 
discusses the limitations of the current framework and outlines potential directions for future work.
\textbf{Appendix~\ref{sec:supp_meteonet_dataset}} 
provides additional experimental results on the MeteoNet dataset.
\textbf{Appendix~\ref{sec:supp_qualitative}} 
provides extended qualitative visualizations to further demonstrate the model's forecasting performance across various extreme precipitation events.

\subsection{Extended Formulation of PixelFlowCast}
\label{supp:Extended Formulation of PixelFlowCast}
In this section, we provide a deeper theoretical and geometric intuition behind the PixelFlowCast framework, 
specifically detailing the auxiliary time step and the multi-step sampling mechanism. 

\subsubsection{Geometric Interpretation and the Role of Auxiliary Time Step $r$}
\label{supp:training_details}

To avoid the optimization collapse caused by directly regressing chaotic vector fields in high-dimensional radar pixel spaces,
PixelFlowCast decouples the prediction and optimization spaces through an auxiliary target time step $r \in [0, t]$.
During training,
rather than directly regressing the exact average velocity over the interval $[t, r]$,
the conditional $x$-prediction model $F_\theta$ directly outputs $\hat{X}_{pred} = F_\theta(Z_t, r, t, H_c)$ to approximate the generalized denoised field.
Geometrically,
the parameter $r$ acts as a trajectory parameter that defines a virtual intercept.
The network constructs a secant line passing through the current state $Z_t$ and the target intermediate state $Z_r$,
extending directly to the clean data manifold at $t=0$.
By converting this explicitly bounded pixel-space prediction back into the velocity space,
we obtain the parameterized average velocity $u_\theta$:
\begin{equation}
    u_\theta(Z_t, r, t, H_c) = \frac{Z_t - \hat{X}_{pred}}{t} \equiv \frac{Z_t - Z_r}{t - r}.
\end{equation}

\subsubsection{Extension to Multi-Step Sampling}
\label{supp:inference_strategy}

The original Pixel Mean Flows (PMF) framework~\cite{lu2026onesteplatentfreeimagegeneration} was primarily introduced and evaluated for one-step image generation (1-NFE).
However,
modeling the highly chaotic and complex spatiotemporal dynamics of extreme precipitation systems 
in a single step often leads to the underestimation of high-threshold meteorological details.
To accurately capture these extreme variations and maintain high-fidelity predictions,
PixelFlowCast extends the one-step PMF paradigm to a robust multi-step sampling approach.
During our multi-step inference phase (e.g.,
$N=10$ steps),
the auxiliary target time step is explicitly set to $r = t - \Delta t$ at each sampling interval.
Under this formulation,
the $x$-prediction model $F_\theta$ outputs a local virtual intercept $\hat{X}_{pred}$ corresponding to the subsequent state $Z_{t-\Delta t}$.
Effectively,
this virtual intercept constructs a trajectory from the current state $Z_t$ towards the target intermediate state $Z_{t-\Delta t}$.
Utilizing this intercept,
we compute the average velocity over the interval $[t, t - \Delta t]$,
yielding $u = (Z_t - \hat{X}_{pred})/t$.
Updating the intermediate state $Z_{curr}$ with this average velocity advances the generation process along the data manifold:
\begin{equation}
    Z_{curr} \leftarrow Z_{curr} - u \cdot \Delta t.
\end{equation}

\subsubsection{Noise-Free Extraction at the Final Step}
\label{supp:noise_free_extraction}

This interval-based multi-step formulation provides the geometric basis for the final extraction strategy detailed in Section 3.4 
of the main text.
While iteratively updating $Z_{curr}$ constructs the evolutionary sequence,
integrating continuous velocity fields across discrete steps can still accumulate residual numerical noise in the high-dimensional pixel space,
particularly as $t$ approaches $0$ during the final $N$-th step,
where the target time step $r$ exactly reaches $0$.
Conceptually,
this indicates that the model prediction no longer points to a noisy intermediate state,
but rather intersects directly with the clean data manifold.
Therefore,
instead of performing a final numerical addition to return the accumulated state $Z_0$,
the framework directly extracts this terminal virtual intercept as the generated residual:
\begin{equation}
    \hat{X}_{res} = \hat{X}_{pred}^{(N)}.
\end{equation}

By extracting this noise-free prediction at the final step,
the proposed latent-free framework balances operational efficiency through an accelerated few-step sampling process while preserving high-fidelity meteorological details.

\subsection{Experimental Settings and Implementation Details}
\label{supp:Experimental Settings and Implementation Details}

In this section, we outline the comprehensive experimental setups, 
including dataset preprocessing, detailed network architectures, and training hyperparameters to ensure full reproducibility.

\subsubsection{Additional Dataset and Preprocessing Details}
\label{supp:datasets}

To ensure full reproducibility, 
we detail the exact data preprocessing and evaluation pipelines. 
From the original 49-frame SEVIR events, 
we extract continuous sequences of length 48 (12 context frames and 36 prediction frames) using a sliding window with a stride of 1. 
The raw SEVIR VIL data, stored as 16-bit integers (0 to 255), 
is spatially downsampled to $128 \times 128$ via interpolation and linearly scaled by a factor of $1/255$ to the $[0, 1]$ range for model training.

For evaluation, 
meteorological metrics including the Critical Success Index (CSI) and Heidke Skill Score (HSS) are computed using six specific intensity thresholds: 
16, 74, 133, 160, 181, and 219. 
To match the normalized output space, 
these thresholds are equivalently scaled by $1/255$ prior to binarization. 
For the multi-scale evaluations ($4 \times 4$ and $16 \times 16$ pooling) mentioned in Section 4.1 of the main text, 
local average-pooling is applied to both predictions and ground truths before thresholding.
This procedure relaxes the strict pixel-to-pixel matching constraint and accommodates slight spatial misalignments. 
Finally, 
inference latency is rigorously measured on a single NVIDIA GeForce RTX 3090 (24GB) GPU. 
To eliminate CUDA initialization overhead and asynchronous execution bias, 
we perform a 5-sample warm-up and enforce strict GPU-CPU synchronization to record the exact wall-clock time for generating the complete 36-frame sequence.

\subsubsection{Deterministic Backbone: SimVP}
\label{sec:supp_simvp}

In the first stage of the PixelFlowCast framework, we employ SimVP~\cite{Gao_2022_CVPR} 
as the deterministic base predictor to capture the macroscopic spatiotemporal evolution trend, denoted as $\hat{X}_{coarse}$. 
We summarize the specific hyperparameter configurations of our instantiated SimVP in Table~\ref{tab:simvp_arch}. 
The model takes the past $T_{in}=12$ frames as input and recursively generates the coarse predictions for the target window.

\begin{table}[htbp]
    \centering
    \small 
    \caption{Detailed architecture and hyperparameters of the SimVP backbone.}
    \label{tab:simvp_arch}
    \begin{tabular*}{\columnwidth}{@{\extracolsep{\fill}} llcc @{}}
        \toprule
        \textbf{Module} & \textbf{Component} & \textbf{Channels ($C$)} & \textbf{Blocks ($N$)} \\
        \midrule
        Encoder & Inception Spatial & 64 & 4 \\
        Translator & Inception Temporal & 256 & 8 \\
        Decoder & Inception Spatial & 64 & 4 \\
        \bottomrule
    \end{tabular*}
\end{table}

\subsubsection{Condition Encoder: KANCondNet}
\label{sec:supp_kancondnet}

To effectively guide the PMF predictor, 
KANCondNet is designed to extract precise multi-scale spatiotemporal conditions $H_c$ from the concatenated past context $X_{past}$ 
and the coarse baseline $\hat{X}_{coarse}$. 
As formulated in Section 3.3 of the main text, KANCondNet strategically replaces traditional fixed activations with learnable B-splines. 
To strike an optimal balance between robust generalization and enhanced nonlinear representation capacity without overfitting 
to high-frequency noise, we exclusively deploy KANResnetBlocks at the deepest spatial resolution, 
while the shallower extraction layers retain standard convolutional ResNetBlocks.

In our implementation, the concatenated input ($T_{in} \times C + T_{window} \times C$ channels) is initially mapped 
to a higher-dimensional space and subsequently downsampled across four hierarchical scales. At each spatial resolution level, 
temporal dynamics are further extracted using a Temporal Cues Unit (TCU)~\cite{li2024scrd}. For the deepest representation level, 
a KANResidualBlock is utilized, where the inner Convolutional KAN (ConvKAN) operates with a specified grid size and spline order. 
The detailed architectural progression, spatial dimensions, and key hyperparameters of KANCondNet are comprehensively summarized in Table~\ref{tab:kancondnet_arch}.

\begin{table}[htbp]
    \centering
    \footnotesize
    \caption{Architectural configurations of KANCondNet. Spatial resolution (Res.) is symmetric ($H=W$).}
    \label{tab:kancondnet_arch}
    \begin{tabular*}{\columnwidth}{@{\extracolsep{\fill}} clccl @{}}
        \toprule
        \textbf{Level} & \textbf{Component} & \textbf{$C_{out}$} & \textbf{Res.} & \textbf{Desc. / Hyperparams} \\
        \midrule
        Input & Conv2d & 64 & 128 & Kernel 3, Pad 1 \\
        \midrule
        \multirow{3}{*}{Level 1} 
        & Resnet & 64 & 128 & Standard ResNet \\
        & TCU & 64 & 128 & Temporal extraction \\
        & Conv2d & 128 & 64 & Stride 2 downsampling \\
        \midrule
        \multirow{3}{*}{Level 2} 
        & Resnet & 128 & 64 & Standard ResNet \\
        & TCU & 128 & 64 & Temporal extraction \\
        & Conv2d & 256 & 32 & Stride 2 downsampling \\
        \midrule
        \multirow{3}{*}{Level 3} 
        & Resnet & 256 & 32 & Standard ResNet \\
        & TCU & 256 & 32 & Temporal extraction \\
        & Conv2d & 512 & 16 & Stride 2 downsampling \\
        \midrule
        \multirow{3}{*}{Level 4} 
        & KANRes & 512 & 16 & Dual-path (Spline+Base); \\
        & & & & grid\_size=3, spline\_order=2 \\
        & TCU & 512 & 16 & Temporal extraction \\
        \bottomrule
    \end{tabular*}
\end{table}

\subsubsection{Pixel Mean Flows Predictor: Modified GTUnet}
\label{sec:supp_gtunet}

\begin{table}[htbp]
    \centering
    \footnotesize
    \caption{Architectural configurations of the Modified GTUnet ($F_\theta$). The base dimension is 64, with channel multipliers $[1, 2, 4, 8]$.}
    \label{tab:gtunet_arch}
    \begin{tabular*}{\columnwidth}{@{\extracolsep{\fill}} clc l @{}}
        \toprule
        \textbf{Level} & \textbf{Component} & \textbf{$C_{out}$} & \textbf{Description \& Blocks} \\
        \midrule
        Input & Modified Conv2d & 64 & Fuses noisy $Z_t$ and condition $H_c$ \\
        \midrule
        \multirow{3}{*}{Down 1} 
        & ResNetBlocks & 64 & 2 Blocks, GroupNorm (groups=8) \\
        & Spatial-Temp Attn & 64 & Extracts local dynamics \\
        & Downsample & 128 & Strided Convolution \\
        \midrule
        \multirow{3}{*}{Down 2} 
        & ResNetBlocks & 128 & 2 Blocks, GroupNorm (groups=8) \\
        & Spatial-Temp Attn & 128 & Extracts local dynamics \\
        & Downsample & 256 & Strided Convolution \\
        \midrule
        \multirow{3}{*}{Down 3} 
        & ResNetBlocks & 256 & 2 Blocks, GroupNorm (groups=8) \\
        & Spatial-Temp Attn & 256 & Extracts local dynamics \\
        & Downsample & 512 & Strided Convolution \\
        \midrule
        \multirow{2}{*}{Middle} 
        & ResNetBlocks & 512 & 2 Blocks, GroupNorm (groups=8) \\
        & Spatial-Temp Attn & 512 & Global context interaction \\
        \midrule
        \multirow{3}{*}{Up 1} 
        & Upsample & 256 & Transposed Convolution \\
        & ResNetBlocks & 256 & 2 Blocks, Concatenation with Down 3 \\
        & Spatial-Temp Attn & 256 & Refines local dynamics \\
        \midrule
        \multirow{3}{*}{Up 2} 
        & Upsample & 128 & Transposed Convolution \\
        & ResNetBlocks & 128 & 2 Blocks, Concatenation with Down 2 \\
        & Spatial-Temp Attn & 128 & Refines local dynamics \\
        \midrule
        \multirow{3}{*}{Up 3} 
        & Upsample & 64 & Transposed Convolution \\
        & ResNetBlocks & 64 & 2 Blocks, Concatenation with Down 1 \\
        & Spatial-Temp Attn & 64 & Refines local dynamics \\
        \midrule
        Output & Conv2d Layer & 1 & Final $x$-prediction ($\hat{X}_{pred}$) \\
        \bottomrule
    \end{tabular*}
\end{table}

In Section 3.4 of the main text, the core of our PMF predictor is abstracted as the $x$-prediction model $F_\theta$.
In our implementation, $F_\theta$ is instantiated based on the Global-Temporal U-Net (GTUnet) architecture 
originally proposed in DiffCast~\cite{Yu_2024_CVPR}. 
GTUnet exhibits strong capabilities in modeling complex meteorological spatiotemporal dependencies 
through its hierarchical encoder-decoder structure and interleaved spatial-temporal attention mechanisms. 
Table~\ref{tab:gtunet_arch} details the specific structural progression and hyperparameter settings (e.g., channel multipliers of $[1, 2, 4, 8]$) utilized in our framework.

While GTUnet was originally designed for standard diffusion models operating on discrete time steps (e.g., $T \in \{1, \dots, 1000\}$), 
our PMF framework is formulated in a continuous time domain where both the current time step $t$ and the auxiliary time step $r$ are strictly bounded within $[0, 1]$. 
Directly feeding such small continuous values into standard Sinusoidal Positional Embeddings fails to 
activate the high-frequency dimensions, causing the network to lose temporal distinguishability. 

To resolve this, our Modified GTUnet introduces a scale-and-normalize temporal injection mechanism. 
First, the continuous variables $t$ and $r$ are dynamically scaled up by a factor of 1000 (\textit{i.e.}, $t_{scaled} = t \times 1000$) 
to align with the canonical frequency domains of standard diffusion embeddings. 
However, this naive scaling significantly amplifies the magnitude of the embedded vectors, 
which inevitably induces gradient explosion when propagated through the subsequent temporal Multi-Layer Perceptrons (MLPs). 
Therefore, after the sinusoidal projection, we apply a rigorous feature re-normalization (dividing by 1000) 
to explicitly bound the variance of the temporal condition embeddings. 
This critical modification ensures stable gradient flow during training 
while perfectly preserving the high-frequency temporal resolution required by our latent-free continuous generative process.

\subsubsection{Training Configurations and Optimization Details}
\label{sec:supp_training_config}

The proposed PixelFlowCast framework is implemented using PyTorch and trained with the Distributed Data Parallel (DDP) strategy 
across multiple NVIDIA GeForce RTX 3090 (24GB) GPUs. 
To accommodate the substantial computational footprint inherent in long-term spatiotemporal sequence forecasting, 
we employ Bfloat16 Mixed Precision (bf16-mixed) during training. This strategy significantly reduces memory consumption and accelerates computation, 
allowing us to maintain a micro batch size of 2 per GPU and yield a global effective batch size of 12 across the distributed environment.

While our framework is entirely capable of being trained from scratch, 
empirically, simultaneously optimizing a deterministic spatiotemporal backbone and a continuous generative flow model from random initialization 
can introduce early-stage gradient instability. 
To accelerate convergence and ensure stable optimization within the highly dynamic velocity space, 
the deterministic backbone (SimVP) is initialized with pre-trained weights. 
Crucially, rather than freezing this backbone as in traditional isolated multi-stage pipelines, 
its weights are actively updated alongside the Modified GTUnet and KANCondNet in an end-to-end manner. 

The comprehensive hyperparameters and optimization strategies are summarized in Table~\ref{tab:training_config}.


\begin{table}[htbp]
    \centering
    \small 
    \renewcommand{\arraystretch}{1.2} 
    \caption{Comprehensive training configurations and hyperparameters for PixelFlowCast.}
    \label{tab:training_config}
    \begin{tabularx}{\columnwidth}{@{} X X @{}}
        \toprule
        \textbf{Hyperparameter} & \textbf{Value / Strategy} \\
        \midrule
        Distributed Strategy & DDP \\
        Precision & bf16-mixed \\
        Batch Size (Micro / Global) & 2 / 12 \\
        Max Epochs & 100 \\
        Optimizer & AdamW \\
        Base Learning Rate & $1 \times 10^{-4}$ \\
        LR Scheduler & Cosine Annealing \\
        Warm-up Phase & 10\% Linear \\
        Objective ($\mathcal{L}_{total}$) & MSE ($\mathcal{L}_{coarse}$) + MSE ($\mathcal{L}_{PMF}$) \\
        \bottomrule
    \end{tabularx}
\end{table}

\subsection{Additional Ablation Studies}
\label{sec:additional_ablation}

In this section, we present supplementary ablation studies to further validate our architectural choices, 
focusing on inference efficiency, optimal sampling configurations, and extraction strategies.

\subsubsection{Ablation Study on Inference Speed}
\label{sec:supp_ablation_speed}

Compared to traditional diffusion paradigms, continuous flow-based models intrinsically benefit from significantly accelerated inference. 
Consequently, a detailed discussion on computational speed is omitted from the main manuscript. 
To comprehensively supplement the architectural evaluations, 
this section explicitly provides a quantitative evaluation of both operational inference time (Time) and macroscopic forecasting accuracy (Overall CSI). 
For a fair and rigorous comparison, 
all Continuous Flow Matching (CFM) baselines evaluated herein consistently maintain the 10-step Euler ODE solver configuration utilized in Section 4.4. 
The comprehensive results across all architectural variants are consolidated in Table~\ref{tab:ablation_efficiency}.

\textbf{Generative Paradigm.} 
Directly applying CFM to the original space (Pixel CFM $128^2$) ensures rapid inference at the cost of severe optimization collapse. 
While introducing Variational Autoencoder (VAE) compression (Latent CFM) partially recovers performance, 
it simultaneously incurs heavy encoding decoding latency. 
As continuous flow models achieve faster inference, 
the latency overhead of latent decoders is no longer negligible. 
This VAE decoding overhead can frequently exceed the computational cost of the entire generator itself. 
Our latent-free PMF paradigm elegantly breaks this dilemma. 
By entirely discarding the computationally expensive VAE, it achieves the highest baseline accuracy with highly competitive speed.

\textbf{KANCondNet Configuration.} 
Heavy conditional encoders like GlobalNet or Full-KAN introduce unacceptable latency overheads. 
Moreover, Full-KAN actually degrades the CSI by overfitting to high-frequency meteorological noise. 
Conversely, completely removing KANs (w/o KAN) yields a fast model but lacks the nonlinear capacity for extreme events. 
By confining KANs strictly to the deepest layer, our Deep-KAN configuration maximizes overall CSI with only a marginal 0.12s computational overhead, 
achieving the optimal Pareto balance for real-time nowcasting.

\begin{table}[htbp]
    \centering
    \caption{Quantitative evaluation of forecasting accuracy and inference time across different architectural variants. Best results are highlighted in \textbf{bold} and second-best are \underline{underlined} within each respective ablation group.}
    \label{tab:ablation_efficiency}
    \begin{tabularx}{\columnwidth}{@{} X c c @{}}
        \toprule
        \textbf{Model Variant} & \textbf{Overall CSI $\uparrow$} & \textbf{Time (s) $\downarrow$} \\
        \midrule
        \multicolumn{3}{c}{\textit{Ablation on Generative Paradigm}} \\
        \midrule
        Latent CFM (VAE $16^2$) & 0.2115 & 0.694 \\
        Latent CFM (VAE $32^2$) & \underline{0.2125} & 0.719 \\
        Pixel CFM ($128^2$)   & 0.1998 & \textbf{0.465} \\
        \midrule
        \multicolumn{3}{c}{\textit{Ablation on KANCondNet}} \\
        \midrule
        GlobalNet        & 0.2139 & 1.060 \\
        Full-KAN         & \underline{0.2204} & 0.733 \\
        w/o KAN          & 0.2177 & \textbf{0.462} \\
        \midrule
        \rowcolor{orange!20} 
        \textbf{PixelFlowCast (Ours)} & \textbf{0.2246} & \underline{0.589} \\
        \bottomrule
    \end{tabularx}
\end{table}

\subsubsection{Ablation Study on Inference Steps}
\label{sec:supp_ablation_steps}

As detailed in Section~\ref{supp:inference_strategy}, while the original PMF framework is conceptualized for one-step generation, 
modeling the highly chaotic dynamics of extreme precipitation systems necessitates a multi-step sampling strategy. 
To determine the optimal configuration, we conduct a comprehensive ablation study on the number of inference steps (1, 5, 10, 15, 20), 
focusing on higher VIL thresholds that represent severe weather events. 

\begin{figure}[htbp]
    \centering
    \includegraphics[width=\linewidth]{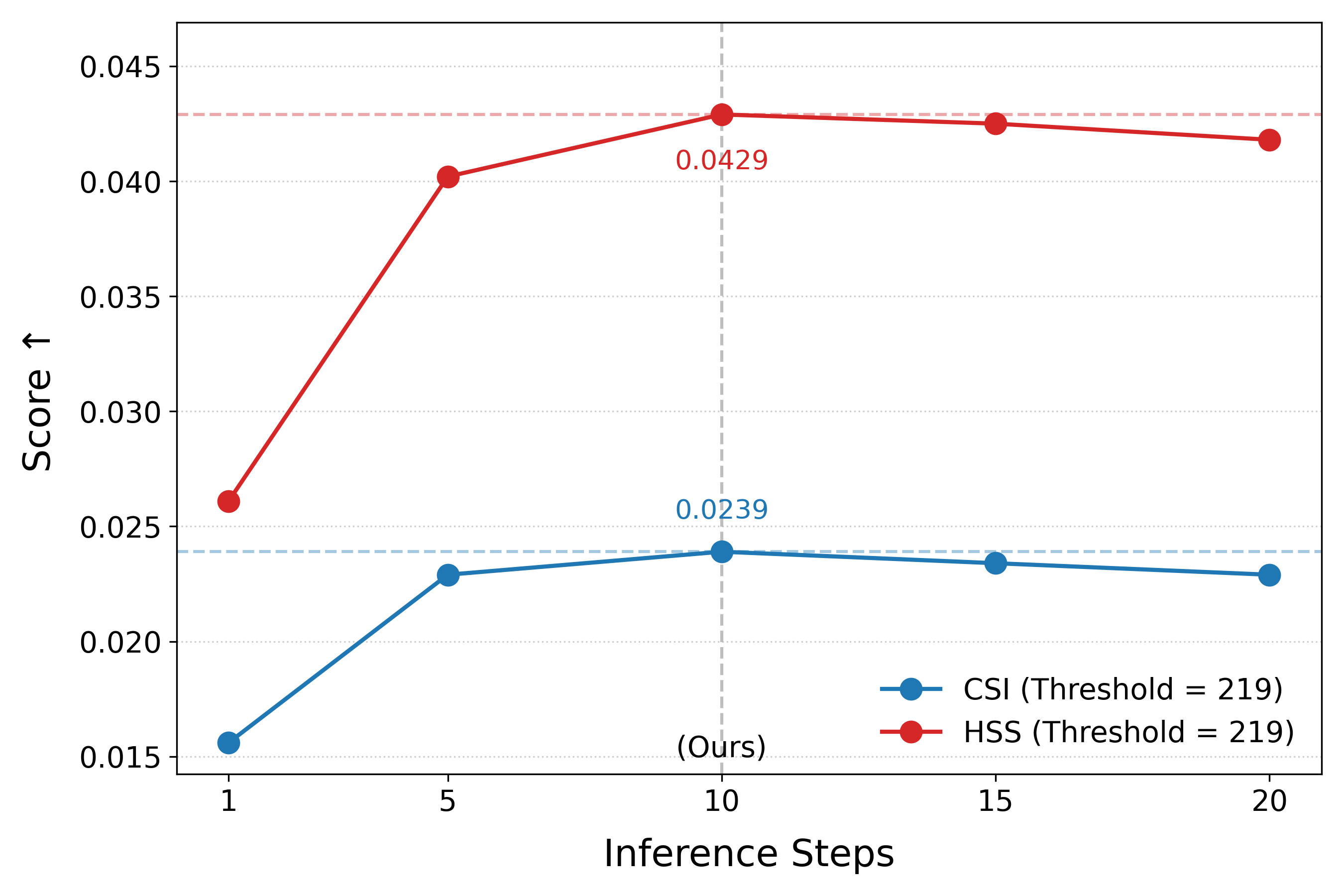}
    \caption{Ablation study of inference steps for CSI and HSS at the extreme precipitation threshold (VIL = 219).}
    \label{fig:steps_ablation_219}
\end{figure}

The quantitative results are summarized in Table~\ref{tab:steps_ablation} and explicitly visualized for the extreme threshold ($T=219$) in Figure~\ref{fig:steps_ablation_219}.
As expected, the 1-step and 5-step configurations struggle to 
faithfully construct the complex evolution trajectories of extreme precipitation, 
yielding sub-optimal performance at high thresholds. 
Increasing the sampling resolution to 10 steps significantly enhances the generation quality, 
achieving the peak CSI and HSS for $T \ge 160$. 

Interestingly, further increasing the steps to 15 or 20 introduces a slight degradation in meteorological performance. 
This phenomenon can be attributed to the accumulation of numerical integration errors across excessive discrete steps in the high-dimensional pixel space. 
Continually updating the current state with parameterized velocity fields over too many intervals inevitably accumulates residual numerical noise. 
This forces the multi-step generation process to over-smooth the pixel-space trajectories, 
thereby washing out the sharp, localized structural details that are critical for predicting extreme precipitation centers.

Regarding efficiency, the 10-step PixelFlowCast ($0.589$ s) remains significantly faster than generative baselines 
like FlowCast ($0.788$ s) and DiffCast ($11.556$ s) (Table 2). 
Thus, it is adopted as the optimal default configuration, achieving a superior Pareto balance between extreme event fidelity and speed.

\begin{table}[htbp]
    \centering
    \small 
    \setlength{\tabcolsep}{3.5pt} 
    \caption{Ablation study of PixelFlowCast performance under different inference sampling steps. Evaluated across higher VIL thresholds on the SEVIR dataset. All metrics are averaged over the full 3-hour forecast horizon (36 frames). Best results are highlighted in \textbf{bold} and second-best are \underline{underlined}.}
    \label{tab:steps_ablation}
    \begin{tabular}{@{}ll cccc c@{}}
        \toprule
        \multirow{2}{*}{\textbf{Metric}} & \multirow{2}{*}{\textbf{Steps}} & \multicolumn{4}{c}{\textbf{VIL Thresholds}} & \multirow{2}{*}{\textbf{Time (s) $\downarrow$}} \\
        \cmidrule(lr){3-6}
        & & \textbf{133} & \textbf{160} & \textbf{181} & \textbf{219} & \\
        \midrule
        \multirow{5}{*}{CSI $\uparrow$} 
        & 1 & 0.1422 & 0.0497 & 0.0346 & 0.0156 & \textbf{0.157} \\
        & 5 & \textbf{0.1552} & 0.0621 & 0.0455 & 0.0229 & \underline{0.306} \\
        & \cellcolor{orange!20}\textbf{10 (Ours)} & \cellcolor{orange!20}\underline{0.1541} & \cellcolor{orange!20}\textbf{0.0636} & \cellcolor{orange!20}\textbf{0.0470} & \cellcolor{orange!20}\textbf{0.0239} & \cellcolor{orange!20}0.589 \\
        & 15 & 0.1522 & \underline{0.0631} & \underline{0.0466} & \underline{0.0234} & 0.694 \\
        & 20 & 0.1503 & 0.0625 & 0.0460 & 0.0229 & 0.888 \\
        \midrule
        \multirow{5}{*}{HSS $\uparrow$} 
        & 1 & 0.2221 & 0.0803 & 0.0552 & 0.0261 & \textbf{0.157} \\
        & 5 & \textbf{0.2439} & 0.1041 & 0.0769 & 0.0402 & \underline{0.306} \\
        & \cellcolor{orange!20}\textbf{10 (Ours)} & \cellcolor{orange!20}\underline{0.2434} & \cellcolor{orange!20}\textbf{0.1079} & \cellcolor{orange!20}\textbf{0.0809} & \cellcolor{orange!20}\textbf{0.0429} & \cellcolor{orange!20}0.589 \\
        & 15 & 0.2412 & \textbf{0.1079} & \textbf{0.0809} & \underline{0.0425} & 0.694 \\
        & 20 & 0.2389 & \underline{0.1072} & \underline{0.0802} & 0.0418 & 0.888 \\
        \bottomrule
    \end{tabular}
\end{table}

\subsubsection{Ablation Study on Noise-Free Extraction Strategy}
\label{sec:supp_ablation_extraction}

In Section~\ref{supp:noise_free_extraction}, 
we proposed a Noise-Free Extraction strategy for the final sampling step. 
Rather than performing a final numerical integration to obtain the accumulated state $Z_{curr}$, 
our framework directly outputs the terminal virtual intercept $\hat{X}_{pred}$ to bypass residual numerical noise. 
To empirically validate this design, we compare the forecasting performance between these two strategies.

As shown in Table~\ref{tab:extraction_strategy}, the Noise-Free Extraction ($\hat{X}_{pred}$) 
demonstrates a consistent advantage across all severe weather thresholds (VIL $\ge 133$). 
Performing numerical integration at the very end of the generation trajectory inevitably introduces slight numerical jitter. 
This residual noise is destructive to the sharp, high-frequency structural details required for predicting extreme precipitation centers. 
By directly extracting the terminal virtual intercept $\hat{X}_{pred}$, 
our strategy bypasses this final accumulation step and successfully preserves the upper bound fidelity of extreme events.

\begin{table}[htbp]
    \centering
    \small
    \setlength{\tabcolsep}{3pt} 
    \caption{Ablation study on the final state extraction strategy. Evaluated on higher VIL thresholds representing extreme precipitation events.}
    \label{tab:extraction_strategy}
    \resizebox{\columnwidth}{!}{
        \begin{tabular}{@{}ll cccc@{}}
            \toprule
            \multirow{2}{*}{\textbf{Metric}} & \multirow{2}{*}{\textbf{Final Extraction Strategy}} & \multicolumn{4}{c}{\textbf{VIL Thresholds}} \\
            \cmidrule(lr){3-6}
            & & \textbf{133} & \textbf{160} & \textbf{181} & \textbf{219} \\
            \midrule
            \multirow{2}{*}{CSI $\uparrow$} 
            & Accumulation ($Z_{curr}$) & 0.1538 & 0.0634 & 0.0468 & 0.0236 \\
            & \cellcolor{orange!20}\textbf{Noise-Free ($\hat{X}_{pred}$, Ours)} & \cellcolor{orange!20}\textbf{0.1541} & \cellcolor{orange!20}\textbf{0.0636} & \cellcolor{orange!20}\textbf{0.0470} & \cellcolor{orange!20}\textbf{0.0239} \\
            \midrule
            \multirow{2}{*}{HSS $\uparrow$} 
            & Accumulation ($Z_{curr}$) & 0.2430 & 0.1077 & 0.0804 & 0.0424 \\
            & \cellcolor{orange!20}\textbf{Noise-Free ($\hat{X}_{pred}$, Ours)} & \cellcolor{orange!20}\textbf{0.2434} & \cellcolor{orange!20}\textbf{0.1079} & \cellcolor{orange!20}\textbf{0.0809} & \cellcolor{orange!20}\textbf{0.0429} \\
            \bottomrule
        \end{tabular}
    }
\end{table}

\subsection{Limitations and Future Work}
\label{sec:Limitations and Future Work}

Despite the effectiveness of KANCondNet in capturing complex nonlinear dynamics, 
our current framework remains purely data-driven. 
A key limitation is the absence of a physics-informed mechanism to constrain the generative process with principled meteorological laws
~\cite{RAISSI2019686, Guen_2020_CVPR, rs18020206}. 
In future work, we aim to encode explicit atmospheric physical laws into the PMF mechanism as prior knowledge. 
By synergizing the strong nonlinear approximation capacity of KANs with physics-based regularization, 
we expect to develop a more robust and physically consistent framework for extreme precipitation nowcasting.

\subsection{Evaluation on Additional MeteoNet Dataset}
\label{sec:supp_meteonet_dataset}

To comprehensively evaluate the robustness and generalization capability 
of the proposed PixelFlowCast across different climatic and geographical conditions, 
we replicated the complete set of experiments from the main text on the MeteoNet dataset~\cite{larvor2020meteonet}. 

\begin{table*}[ht]
    \centering
    \caption{Quantitative comparison of PixelFlowCast with baseline models on the MeteoNet dataset. Consistent with our main findings, PixelFlowCast effectively handles highly localized meteorological dynamics, outperforming the best deterministic baseline (SimVP) by 7.8\% in overall CSI while achieving superior visual fidelity (LPIPS of 0.2343). ``Overall Average'' represents metrics computed over the full 3-hour forecast (36 frames), while ``Last 1 Hour'' focuses exclusively on the 2-3 hour prediction phase. Best results are highlighted in \textbf{bold} and second-best are \underline{underlined}.}
    \label{tab:sota_comparison_meteonet}
    \begin{tabular*}{\textwidth}{@{\extracolsep{\fill}} l cccccc cc @{}}
        \toprule
        & \multicolumn{6}{c}{Overall Average (All Thresholds, 3 Hours)} & \multicolumn{2}{c}{Last 1 Hour} \\
        \cmidrule(lr){2-7} \cmidrule(lr){8-9}
        Model & CSI $\uparrow$ & CSI-pool4 $\uparrow$ & CSI-pool16 $\uparrow$ & HSS $\uparrow$ & LPIPS $\downarrow$ & SSIM $\uparrow$ & CSI $\uparrow$ & HSS $\uparrow$ \\
        \midrule
        U-Net~\cite{NEURIPS2020_fa78a161}       & 0.2209 & 0.2783 & 0.3319 & 0.3100 & 0.3675 & 0.6509 & 0.1026 & 0.1562 \\
        Earthformer~\cite{NEURIPS2022_a2affd71} & 0.2398 & 0.2968 & 0.3393 & 0.3370 & 0.4130 & 0.6363 & 0.1309 & 0.1985 \\
        SimVP~\cite{Gao_2022_CVPR}       & \underline{0.2598} & \underline{0.3251} & \underline{0.3562} & \underline{0.3589} & 0.3444 & 0.6813 & 0.1290 & 0.1948 \\
        \midrule
        PreDiff~\cite{NEURIPS2023_f82ba6a6}     & 0.1454 & 0.1357 & 0.1033 & 0.1966 & 0.4044 & 0.5557 & 0.0526 & 0.0898 \\
        FlowCast~\cite{ribeiro2026flowcastadvancingprecipitationnowcasting}    & 0.1997 & 0.2328 & 0.2741 & 0.2850 & 0.3223 & \underline{0.6989} & 0.1029 & 0.1528 \\
        DiffCast~\cite{Yu_2024_CVPR}    & 0.2511 & 0.3076 & 0.3379 & 0.3572 & \underline{0.2926} & 0.6663 & \underline{0.1580} & \underline{0.2368} \\
        \midrule
        \textbf{PixelFlowCast (Ours)} & \textbf{0.2801} & \textbf{0.3378} & \textbf{0.3660} & \textbf{0.3919} & \textbf{0.2343} & \textbf{0.7222} & \textbf{0.1724} & \textbf{0.2583} \\
        \bottomrule
    \end{tabular*}
\end{table*}

\begin{table*}[ht]
    \centering
    \caption{Detailed performance comparison across different precipitation thresholds on the MeteoNet dataset. Consistent with our main analysis, PixelFlowCast robustly captures extreme precipitation dynamics to achieve the highest CSI and HSS across all thresholds, while resolving the fidelity-efficiency trade-off with the fastest inference speed (0.4912s per sample) among generative baselines. All metrics are averaged over the full 3-hour forecast horizon (36 frames), with best results highlighted in \textbf{bold} and second-best \underline{underlined}.}
    \label{tab:threshold_comparison_meteonet}
    \begin{tabular*}{\textwidth}{@{\extracolsep{\fill}} ll cccc c @{}}
        \toprule
        \multirow{2}{*}{\textbf{Metric}} & \multirow{2}{*}{\textbf{Model}} & \multicolumn{4}{c}{\textbf{Precipitation Thresholds}} & \multirow{2}{*}{\textbf{Time (s) $\downarrow$}} \\
        \cmidrule(lr){3-6}
        & & \textbf{12} & \textbf{18} & \textbf{24} & \textbf{32} & \\
        \midrule
        \multirow{4}{*}{CSI $\uparrow$}
        & PreDiff~\cite{NEURIPS2023_f82ba6a6} & 0.2262 & 0.1819 & 0.1273 & 0.0463 & 94.4415 \\
        & FlowCast~\cite{ribeiro2026flowcastadvancingprecipitationnowcasting} & 0.3121 & 0.2512 & 0.1732 & 0.0625 & \underline{0.7670} \\
        & DiffCast~\cite{Yu_2024_CVPR} & \underline{0.3680} & \underline{0.3072} & \underline{0.2269} & \underline{0.1024} & 9.9112 \\
        & \textbf{PixelFlowCast (Ours)} & \textbf{0.4044} & \textbf{0.3441} & \textbf{0.2568} & \textbf{0.1152} & \textbf{0.4912} \\
        \midrule
        \multirow{4}{*}{HSS $\uparrow$}
        & PreDiff~\cite{NEURIPS2023_f82ba6a6} & 0.2876 & 0.2439 & 0.1815 & 0.0733 & 94.4415 \\
        & FlowCast~\cite{ribeiro2026flowcastadvancingprecipitationnowcasting} & 0.4182 & 0.3547 & 0.2615 & 0.1057 & \underline{0.7670} \\
        & DiffCast~\cite{Yu_2024_CVPR} & \underline{0.4886} & \underline{0.4291} & \underline{0.3389} & \underline{0.1721} & 9.9112 \\
        & \textbf{PixelFlowCast (Ours)} & \textbf{0.5291} & \textbf{0.4717} & \textbf{0.3765} & \textbf{0.1905} & \textbf{0.4912} \\
        \bottomrule
    \end{tabular*}
\vspace{-3mm}
\end{table*}
\subsubsection{Dataset.} 
The MeteoNet dataset is provided by the French national meteorological service (Météo-France). 
This dataset captures the evolution of radar echoes over the French territory, 
featuring a high spatial resolution of $0.01^\circ$ (approximately 1 km) on an original grid of $565 \times 784$ pixels, 
and a temporal resolution of 5 minutes. 
Following a consistent experimental protocol, we use a chronological split to divide the events 
from the 2016--2018 observation period into independent training, validation, and test sets. 
The training set covers the period from January 2016 to December 2017, containing 4,607 sequence samples; 
the validation set spans from January 2018 to May 2018, with 1,449 sequence samples; 
and the test set is from June 2018 to December 2018, totaling 589 sequence samples.

\subsubsection{Data Preprocessing.} 
Consistent with the preprocessing strategy applied to the SEVIR dataset, 
we addressed computing resource limitations by cropping and downsampling the spatial dimensions 
of all original MeteoNet radar frames to $128 \times 128$ pixels during the preprocessing stage. 
In the construction of the time series, 
each sample is a 48-frame sequence extracted from the continuous radar observations. 
The model receives the radar observation fields from the past 1 hour (i.e., 12 consecutive frames) as the input context, 
aiming to forecast the spatiotemporal evolution of the precipitation system for the next 3 hours (i.e., 36 consecutive frames).
The data range of the MeteoNet radar frames is set to [0, 70] dBZ, 
and the CSI and HSS are computed at specific precipitation thresholds (12, 18, 24, 32), 
following the standard evaluation protocol~\cite{larvor2020meteonet}.

\subsubsection{Quantitative Results}
\begin{figure*}[ht] 
    \centering
    
    \begin{minipage}[b]{0.48\textwidth}
        \centering
        \includegraphics[trim=10 5 10 5, clip, width=0.9\linewidth]{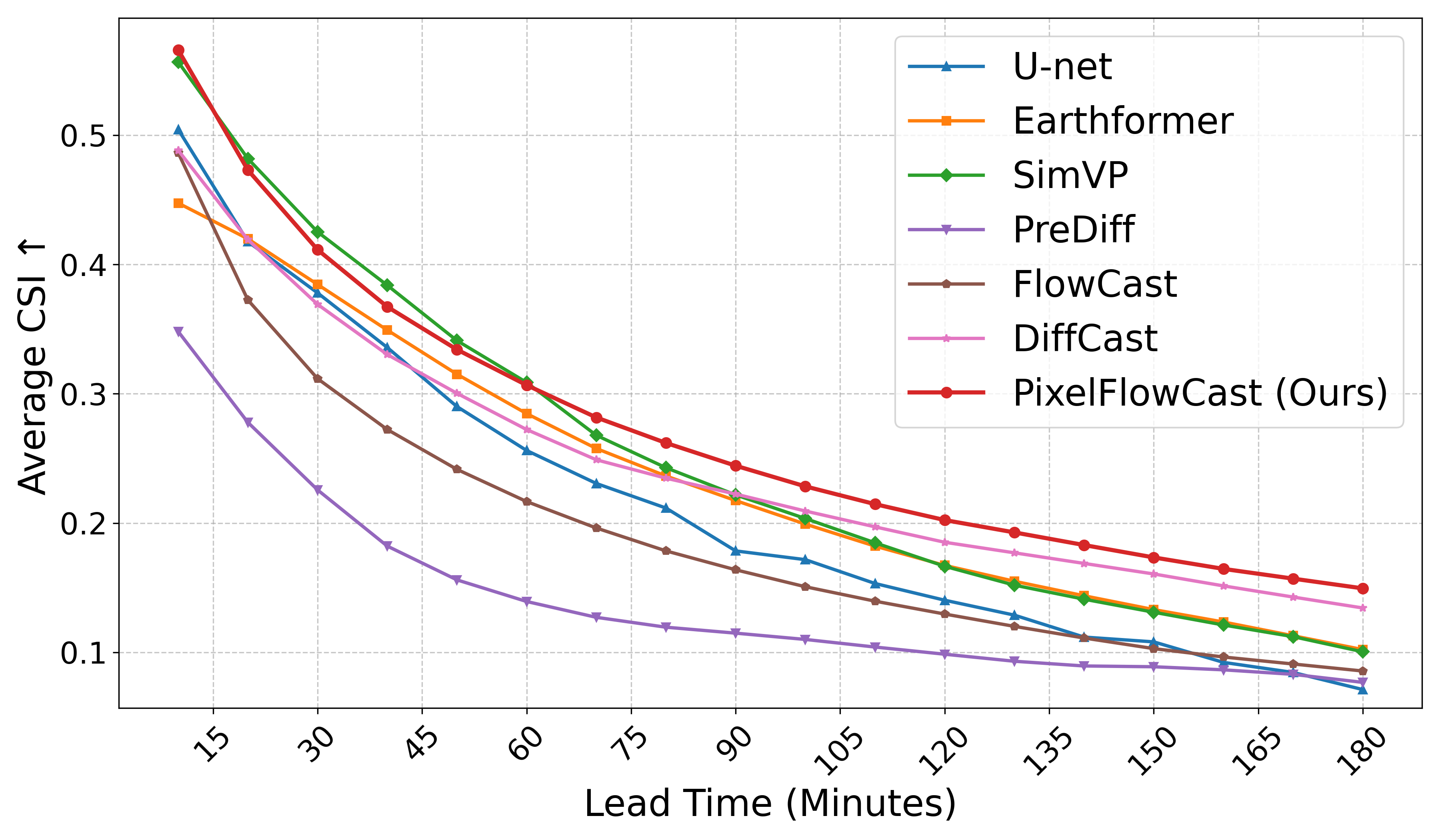}
        \caption{Average CSI degradation over a 3-hour forecast lead time on the MeteoNet dataset. Consistent with our main findings, PixelFlowCast effectively mitigates error accumulation and consistently suppresses performance decay across the entire 180-minute horizon.}
        \label{fig:temporal_csi_meteonet}
    \end{minipage}
    \hfill 
    \begin{minipage}[b]{0.48\textwidth}
        \centering
        \includegraphics[trim=10 5 10 5, clip, width=0.9\linewidth]{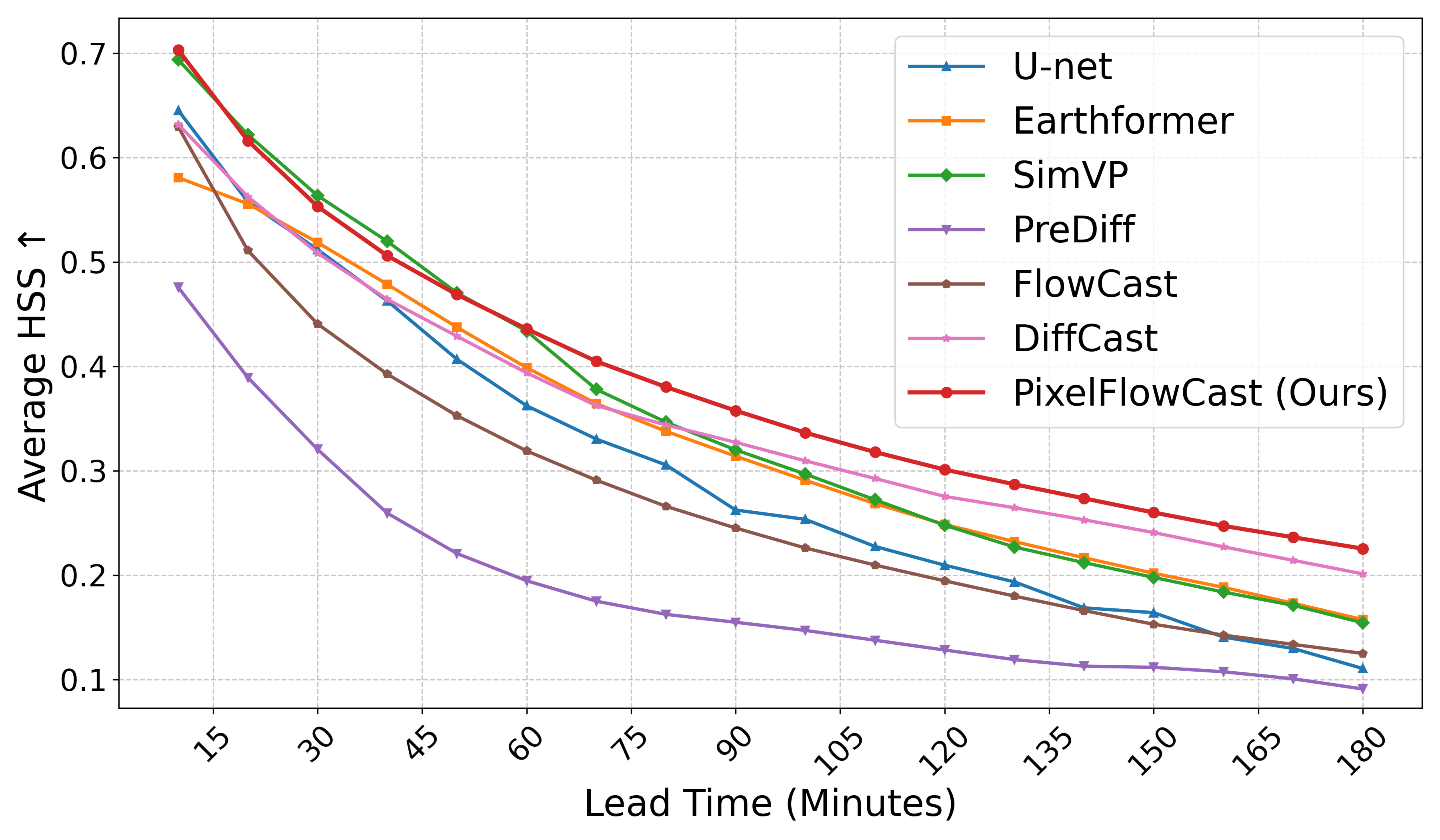}
        \caption{Average CSI degradation over a 3-hour forecast lead time on the MeteoNet dataset. Consistent with our main findings, PixelFlowCast effectively mitigates error accumulation and consistently suppresses performance decay across the entire 180-minute horizon.}
        \label{fig:temporal_hss_meteonet}
    \end{minipage}

\end{figure*}

\begin{figure*}[ht] 
    \centering
    
    \begin{minipage}[b]{0.48\textwidth}
        \centering
        \includegraphics[trim=1.2cm 1.5cm 1.5cm 1.5cm, clip, width=0.75\linewidth]{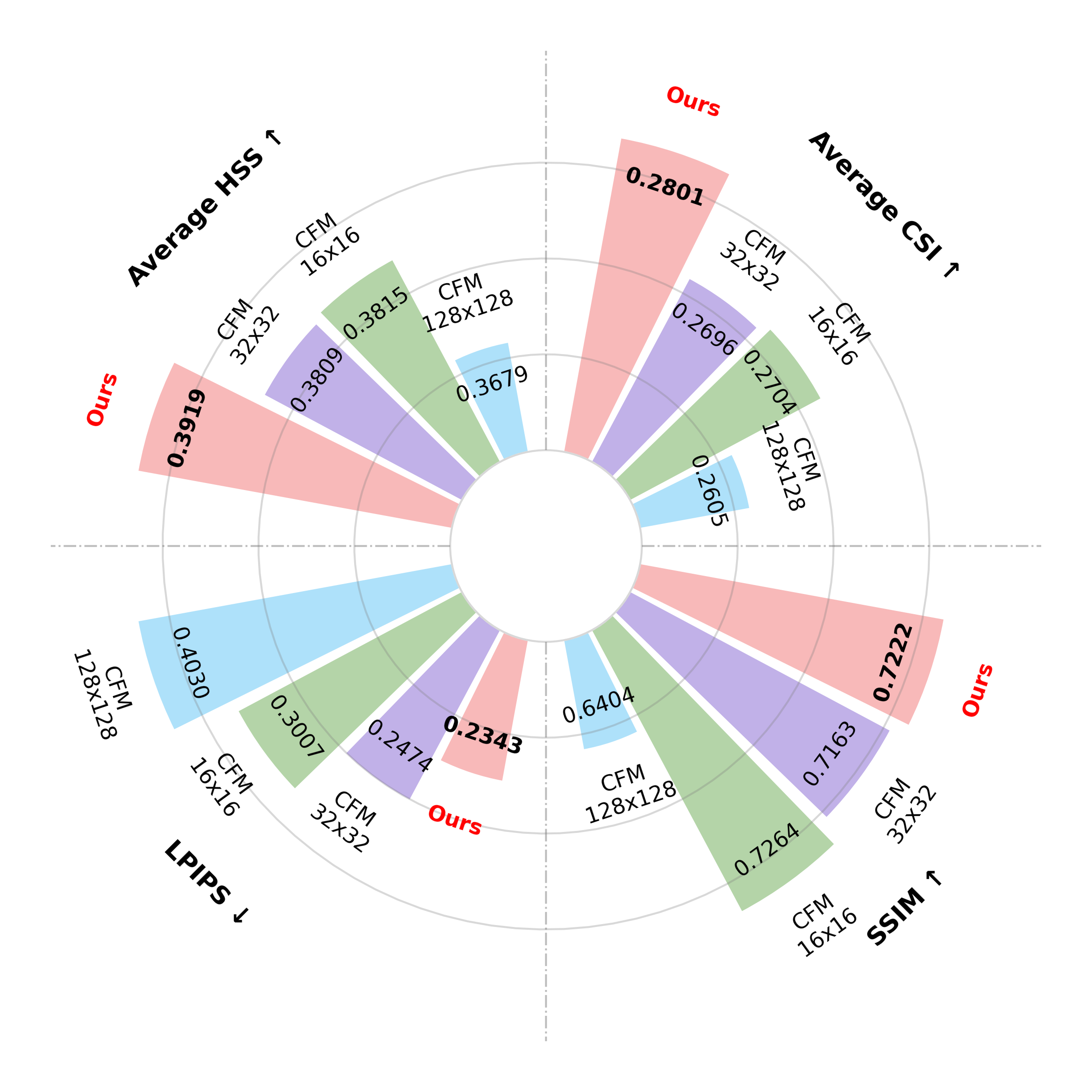}
        \caption{Ablation study on PMF generative paradigm on the MeteoNet dataset. Results confirm that discarding latent compression in favor of direct pixel-space residual prediction effectively avoids the smoothing effects of VAEs, yielding optimal accuracy and visual fidelity.}
        \label{fig:ablation_pmf_vs_fm_meteonet}
    \end{minipage}
    \hfill 
    \begin{minipage}[b]{0.48\textwidth}
        \centering
        \includegraphics[trim=1.1cm 1.1cm 1.1cm 1.1cm, clip, width=0.75\linewidth]{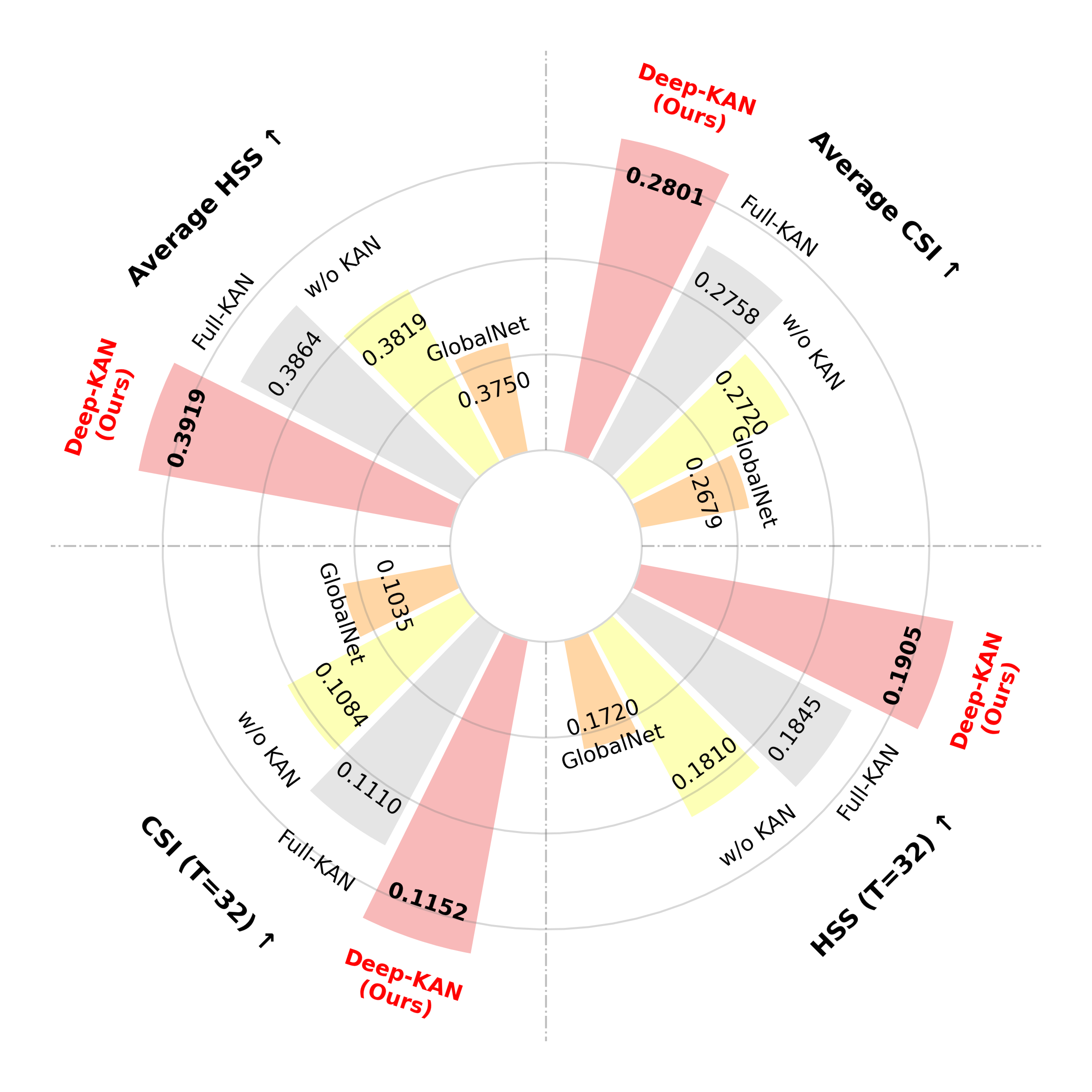}
        \caption{Ablation study of KANCondNet module on the MeteoNet dataset. Confining KANResNetBlocks strictly to the deepest layer (Deep-KAN) achieves the best balance between nonlinear expressiveness and robust generalization, effectively preventing overfitting on chaotic noise.}
        \label{fig:ablation_kancondnet_meteonet}
    \end{minipage}

\end{figure*}

The quantitative and qualitative results on the MeteoNet dataset are summarized 
in Tables~\ref{tab:sota_comparison_meteonet} and \ref{tab:threshold_comparison_meteonet}, 
as well as Figures~\ref{fig:temporal_csi_meteonet}, \ref{fig:temporal_hss_meteonet}, 
\ref{fig:ablation_pmf_vs_fm_meteonet}, \ref{fig:ablation_kancondnet_meteonet}. 
Overall, the empirical performance exhibits a highly consistent trend with those observed on the SEVIR dataset, 
further validating the effectiveness and generalizability of PixelFlowCast.




\subsection{More Qualitative Results}
\label{sec:supp_qualitative}

\begin{figure*}[p]
    \begin{minipage}[c][\textheight][c]{\textwidth}
        \centering
        \includegraphics[width=\textwidth]{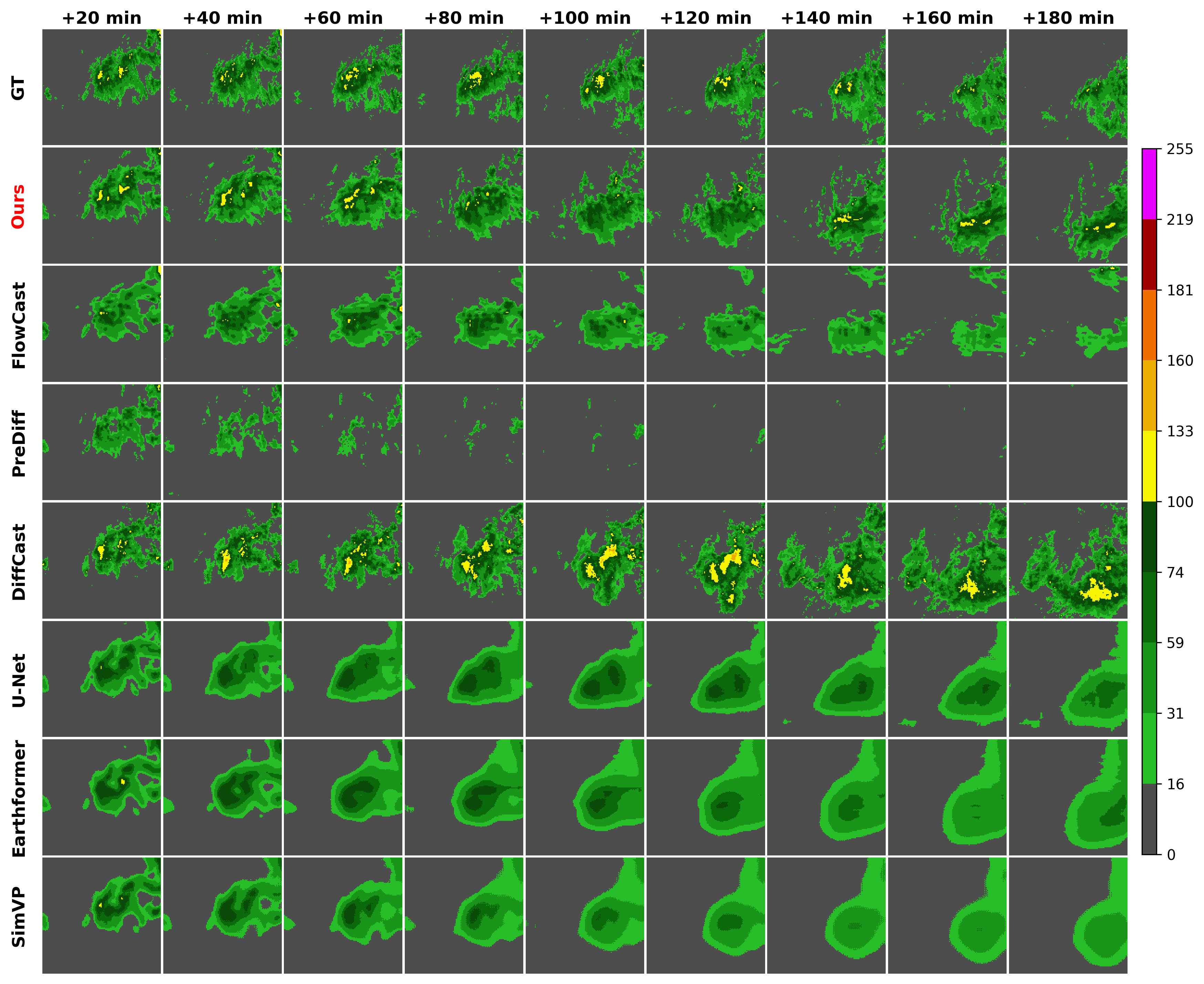}
        \caption{Additional qualitative comparison (Example 1). By operating directly in the pixel space, PixelFlowCast preserves fine radar textures and sharp boundaries over the entire 180-minute forecast horizon, effectively avoiding the progressive blurring characteristic of traditional models.}
        \label{fig:supp_example1}
    \end{minipage}
\end{figure*}

\begin{figure*}[p]
    \begin{minipage}[c][\textheight][c]{\textwidth}
        \centering
        \includegraphics[width=\textwidth]{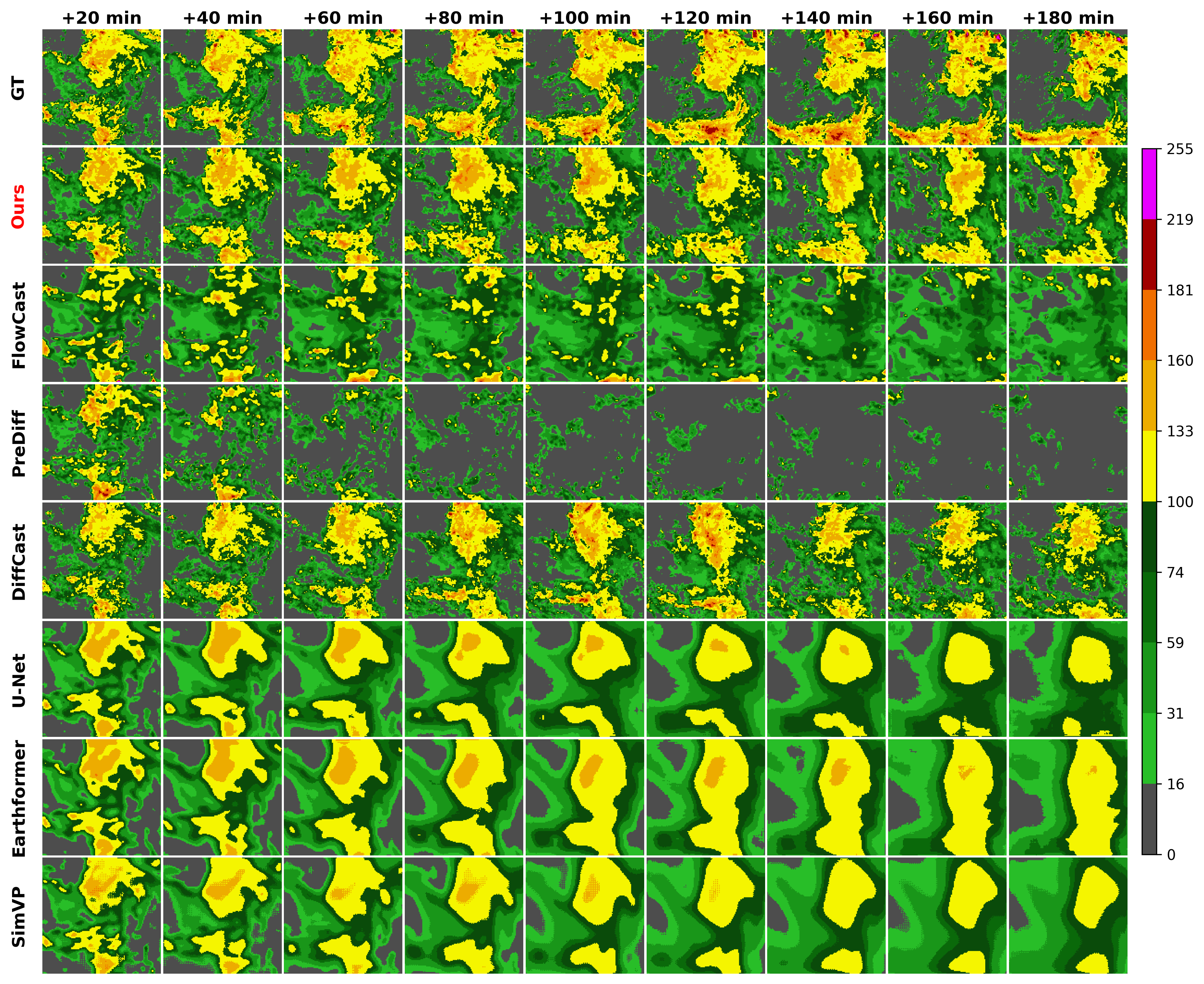}
        \caption{Additional qualitative comparison (Example 2). PixelFlowCast demonstrates robust long-term forecasting capability, accurately capturing the spatiotemporal evolution of extreme precipitation events while maintaining high-intensity precipitation cores without structural decay.}
        \label{fig:supp_example2}
    \end{minipage}
\end{figure*}

\begin{figure*}[p]
    \begin{minipage}[c][\textheight][c]{\textwidth}
        \centering
        \includegraphics[width=\textwidth]{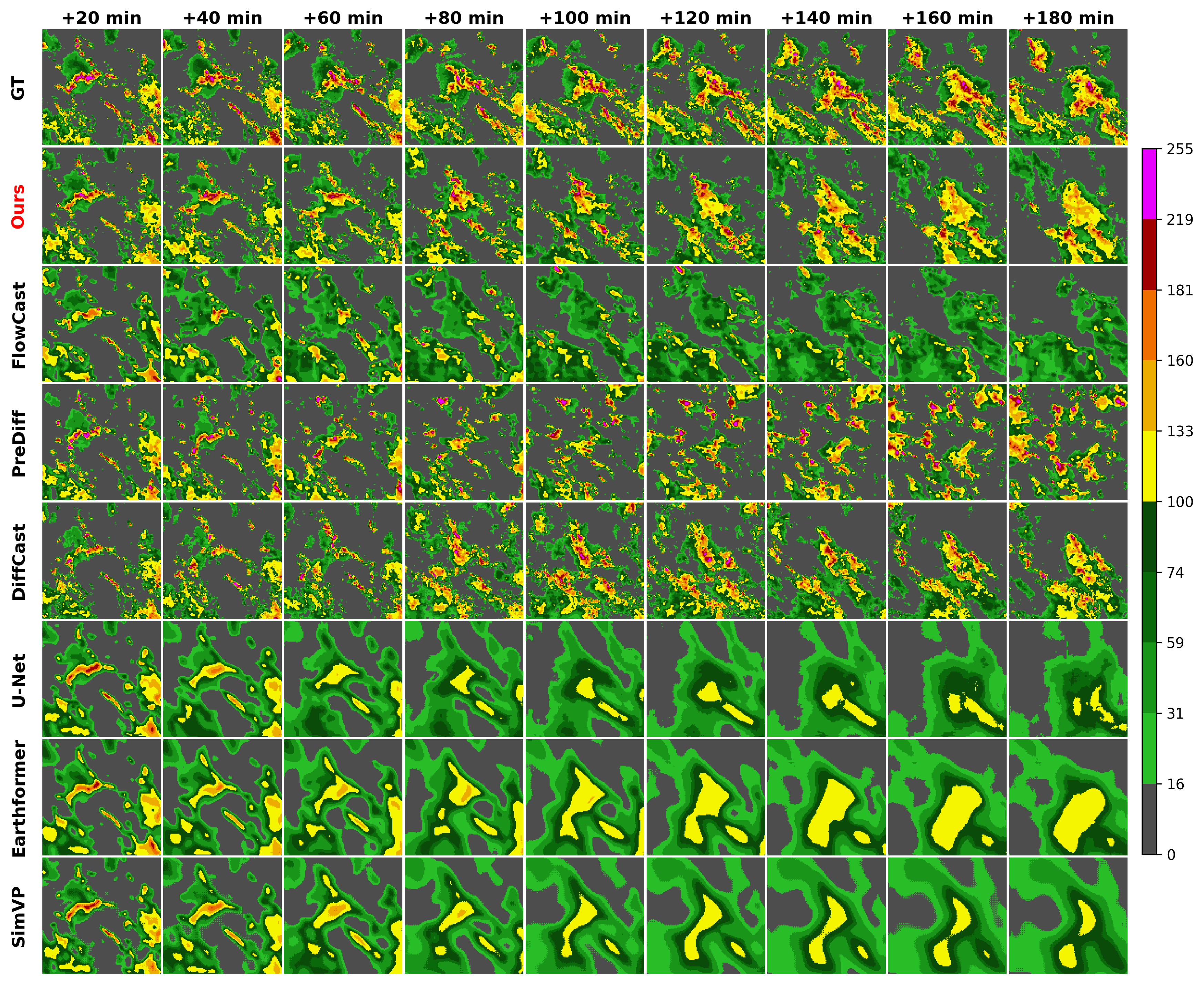}
        \caption{Additional qualitative comparison (Example 3). Consistent with previous examples, our model successfully resolves microscopic physical details and localized heavy precipitation areas, which are often smoothed out by methods relying on latent space compression.}
        \label{fig:supp_example3}
    \end{minipage}
\end{figure*}

\begin{figure*}[p]
    \begin{minipage}[c][\textheight][c]{\textwidth}
        \centering
        \includegraphics[width=\textwidth]{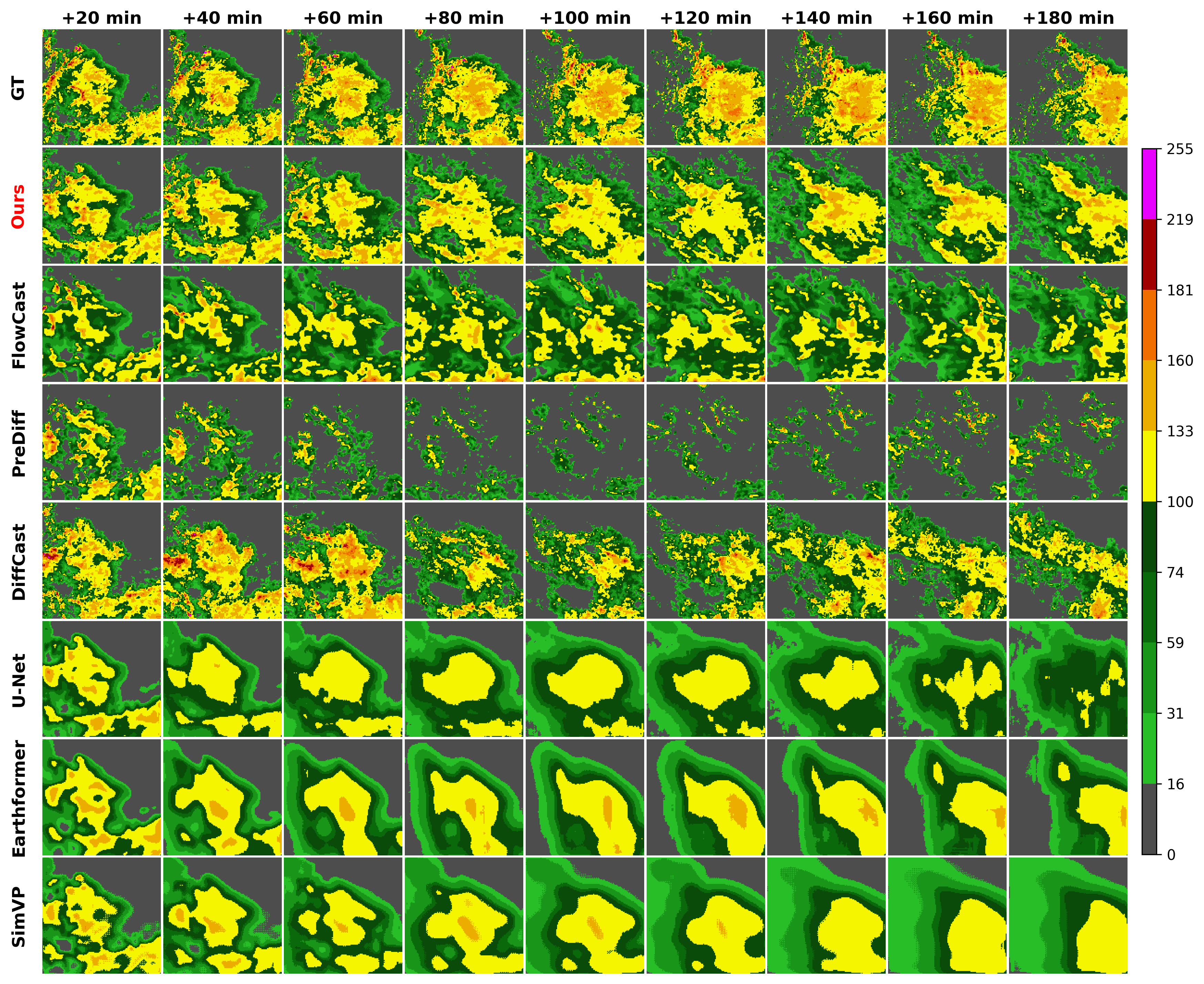}
        \caption{Additional qualitative comparison (Example 4). The predictions demonstrate that PixelFlowCast maintains physically realistic spatial distributions and high-fidelity textures even at the later lead times (e.g., up to 180 minutes).}
        \label{fig:supp_example4}
    \end{minipage}
\end{figure*}

\begin{figure*}[p]
    \begin{minipage}[c][\textheight][c]{\textwidth}
        \centering
        \includegraphics[width=\textwidth]{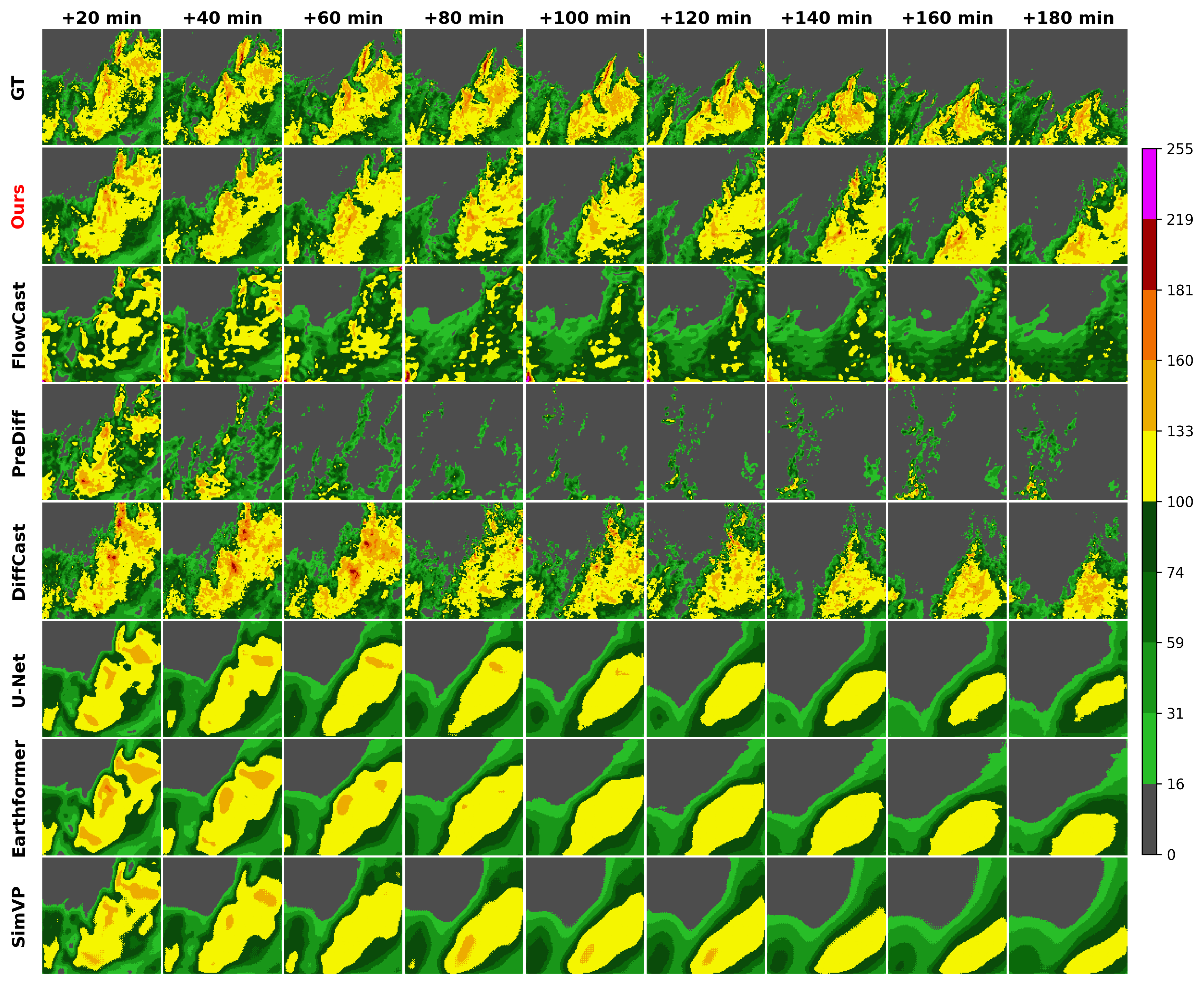}
        \caption{Additional qualitative comparison (Example 5). Further visualization confirming that the proposed framework achieves superior generation quality, preserving critical meteorological details across complex and rapidly changing radar echo sequences.}
        \label{fig:supp_example5}
    \end{minipage}
\end{figure*}

\begin{figure*}[p]
    \begin{minipage}[c][\textheight][c]{\textwidth}
        \centering
        \includegraphics[width=\textwidth]{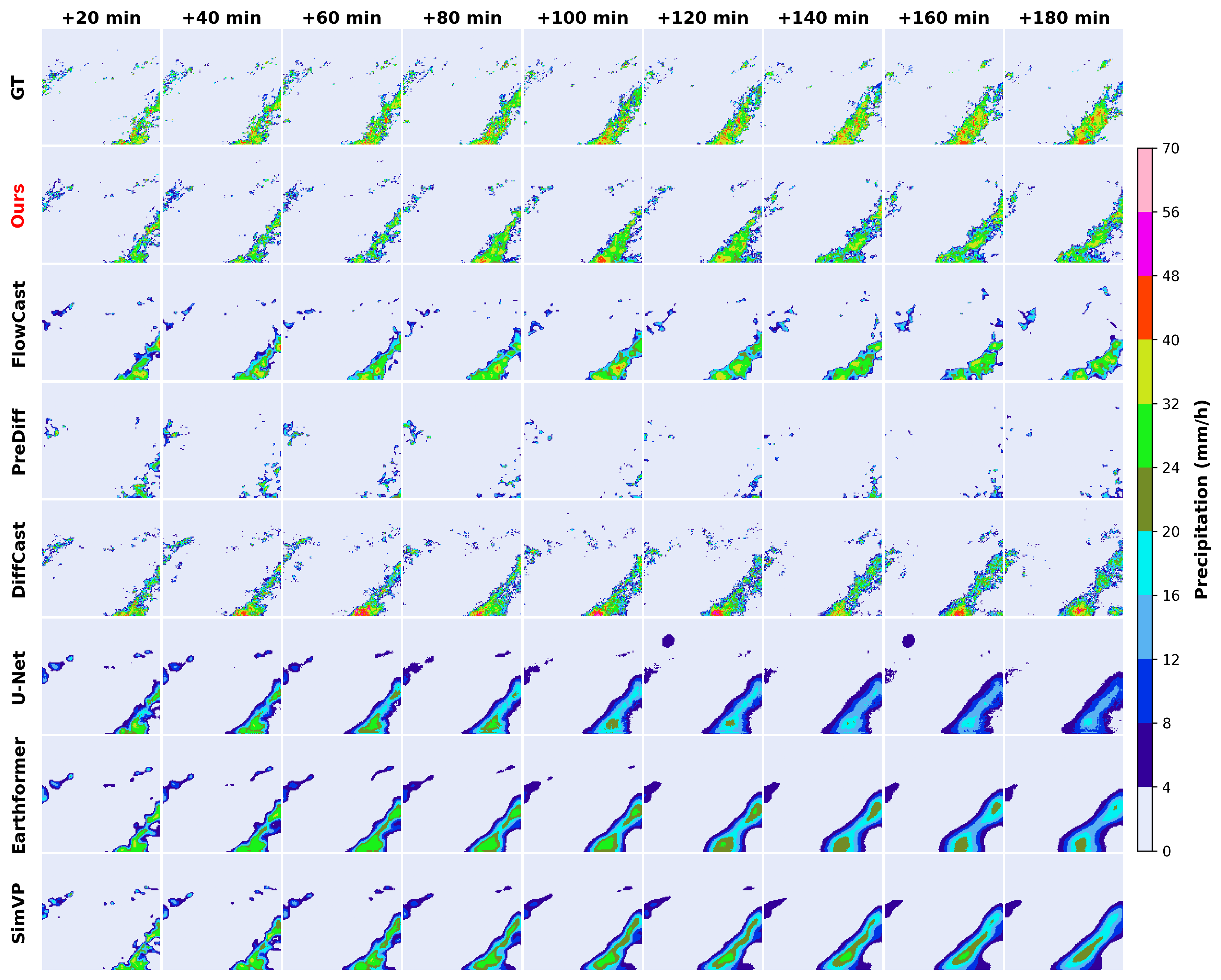}
        \caption{Additional qualitative comparison on the MeteoNet dataset (Example 1). By operating directly in the pixel space, PixelFlowCast preserves fine precipitation textures and sharp boundaries over the entire 180-minute forecast horizon, effectively avoiding the progressive blurring characteristic of traditional deterministic models.}
        \label{fig:meteo_supp_example1}
    \end{minipage}
\end{figure*}

\begin{figure*}[p]
    \begin{minipage}[c][\textheight][c]{\textwidth}
        \centering
        \includegraphics[width=\textwidth]{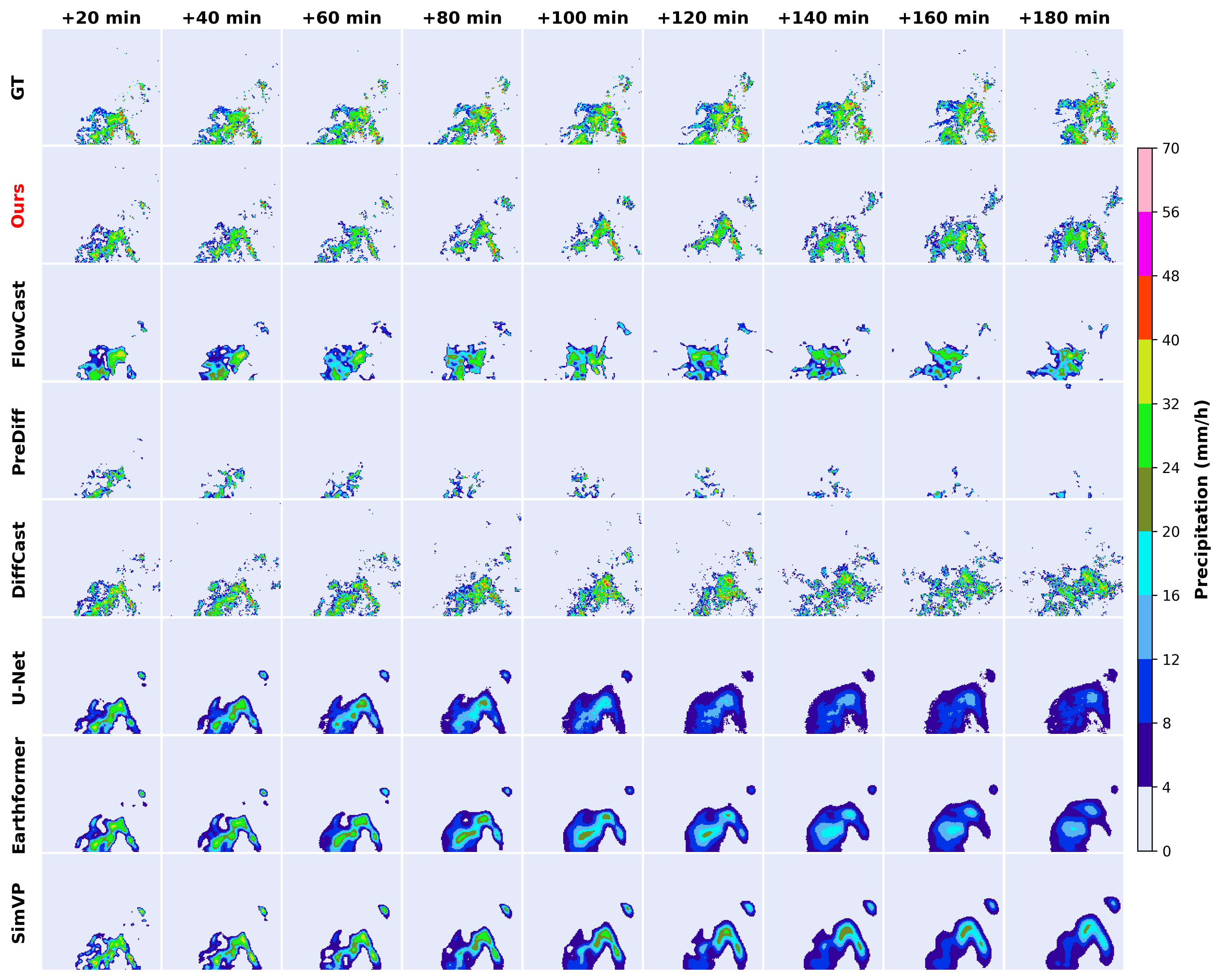}
        \caption{Additional qualitative comparison on the MeteoNet dataset (Example 2). PixelFlowCast demonstrates robust long-term forecasting capability, accurately capturing the spatiotemporal evolution of heavy precipitation bands while maintaining high-intensity precipitation cores without severe structural decay.}
        \label{fig:meteo_supp_example2}
    \end{minipage}
\end{figure*}

\begin{figure*}[p]
    \begin{minipage}[c][\textheight][c]{\textwidth}
        \centering
        \includegraphics[width=\textwidth]{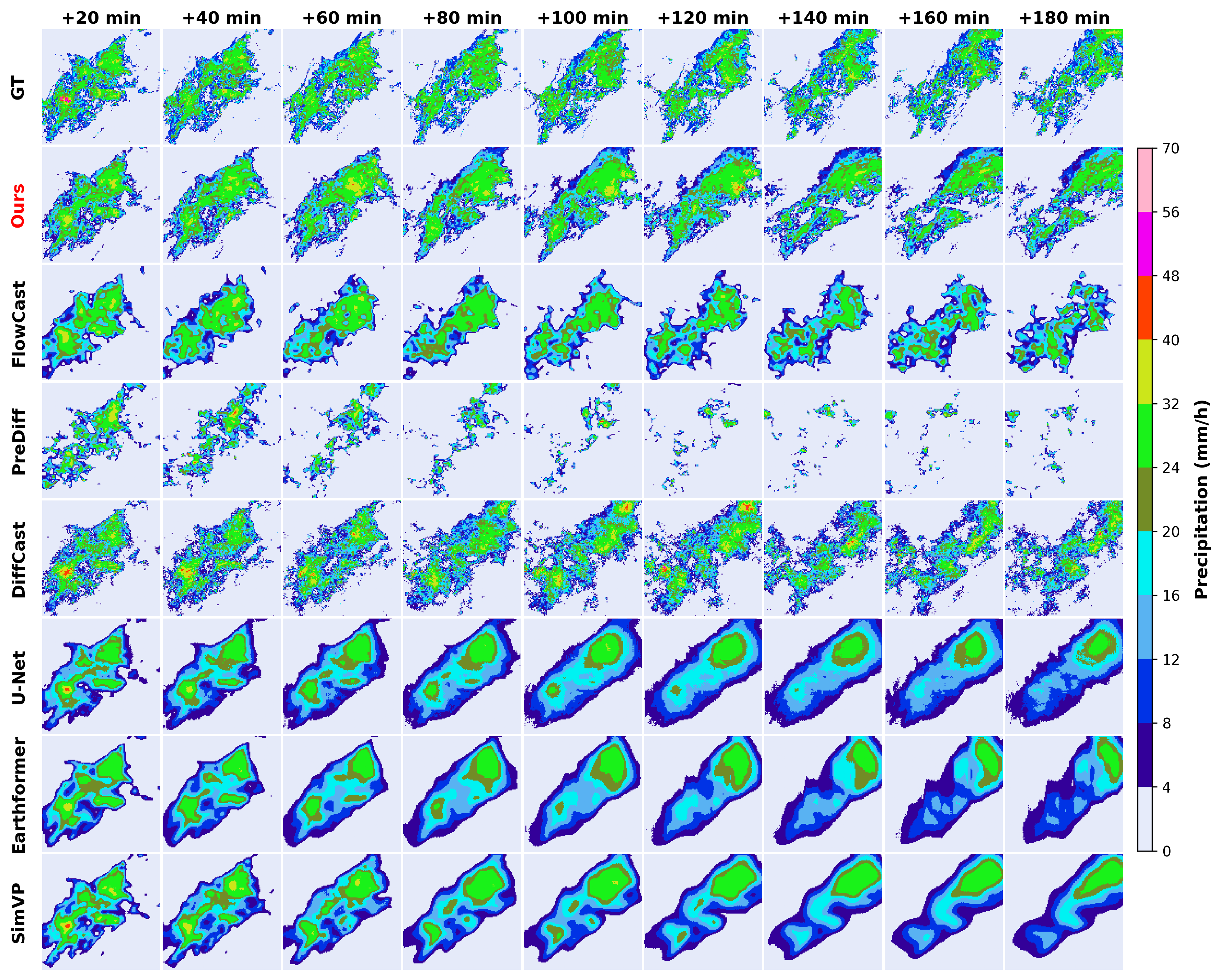}
        \caption{Additional qualitative comparison on the MeteoNet dataset (Example 3). Consistent with previous analyses, our model successfully resolves highly localized meteorological details and intense rainfall areas, which are typically smoothed out by methods relying on latent space compression.}
        \label{fig:meteo_supp_example3}
    \end{minipage}
\end{figure*}

\begin{figure*}[p]
    \begin{minipage}[c][\textheight][c]{\textwidth}
        \centering
        \includegraphics[width=\textwidth]{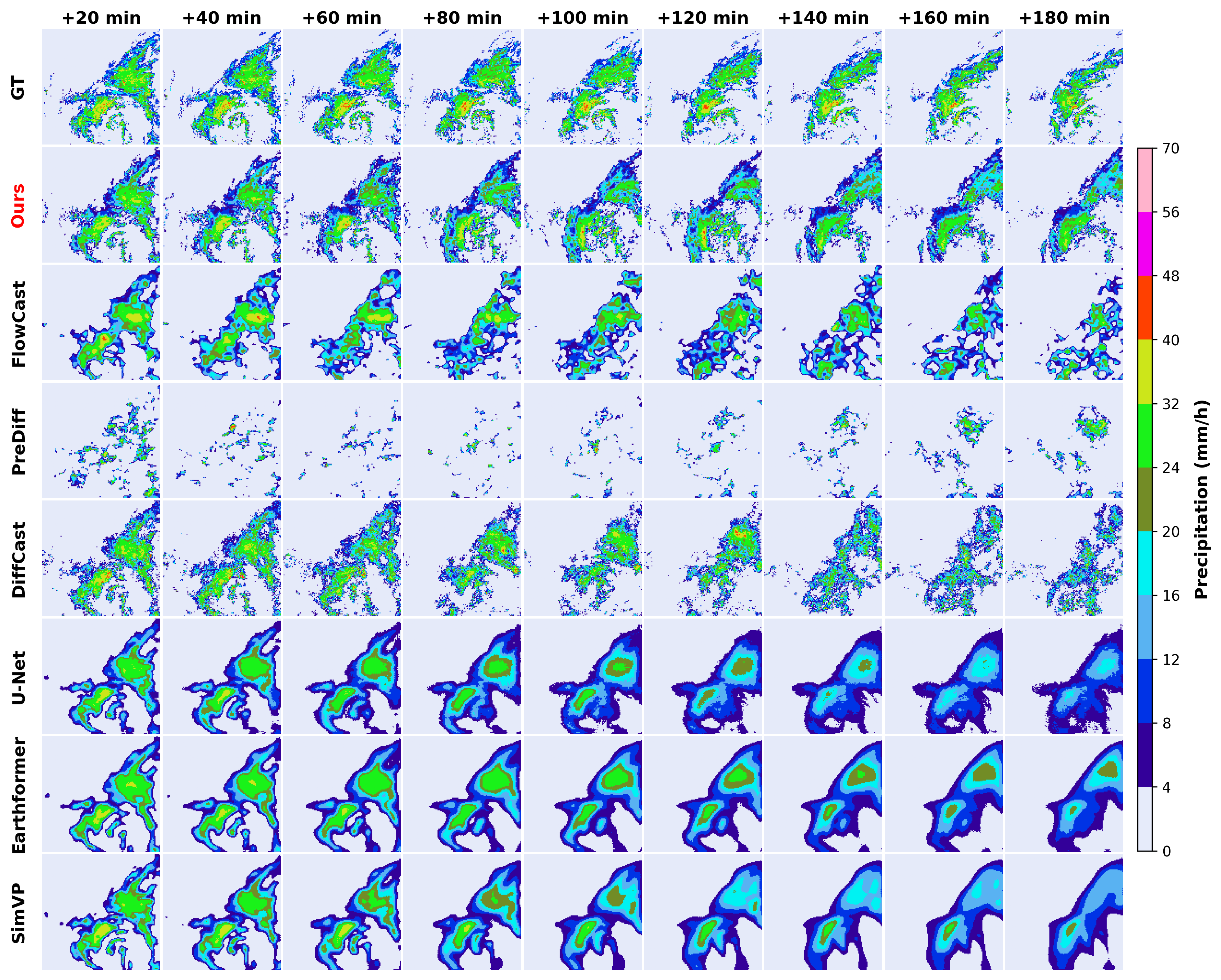}
        \caption{Additional qualitative comparison on the MeteoNet dataset (Example 4). The visualizations demonstrate that PixelFlowCast maintains physically realistic spatial distributions and high-fidelity structures even at the later prediction lead times (up to 180 minutes).}
        \label{fig:meteo_supp_example4}
    \end{minipage}
\end{figure*}

\begin{figure*}[p]
    \begin{minipage}[c][\textheight][c]{\textwidth}
        \centering
        \includegraphics[width=\textwidth]{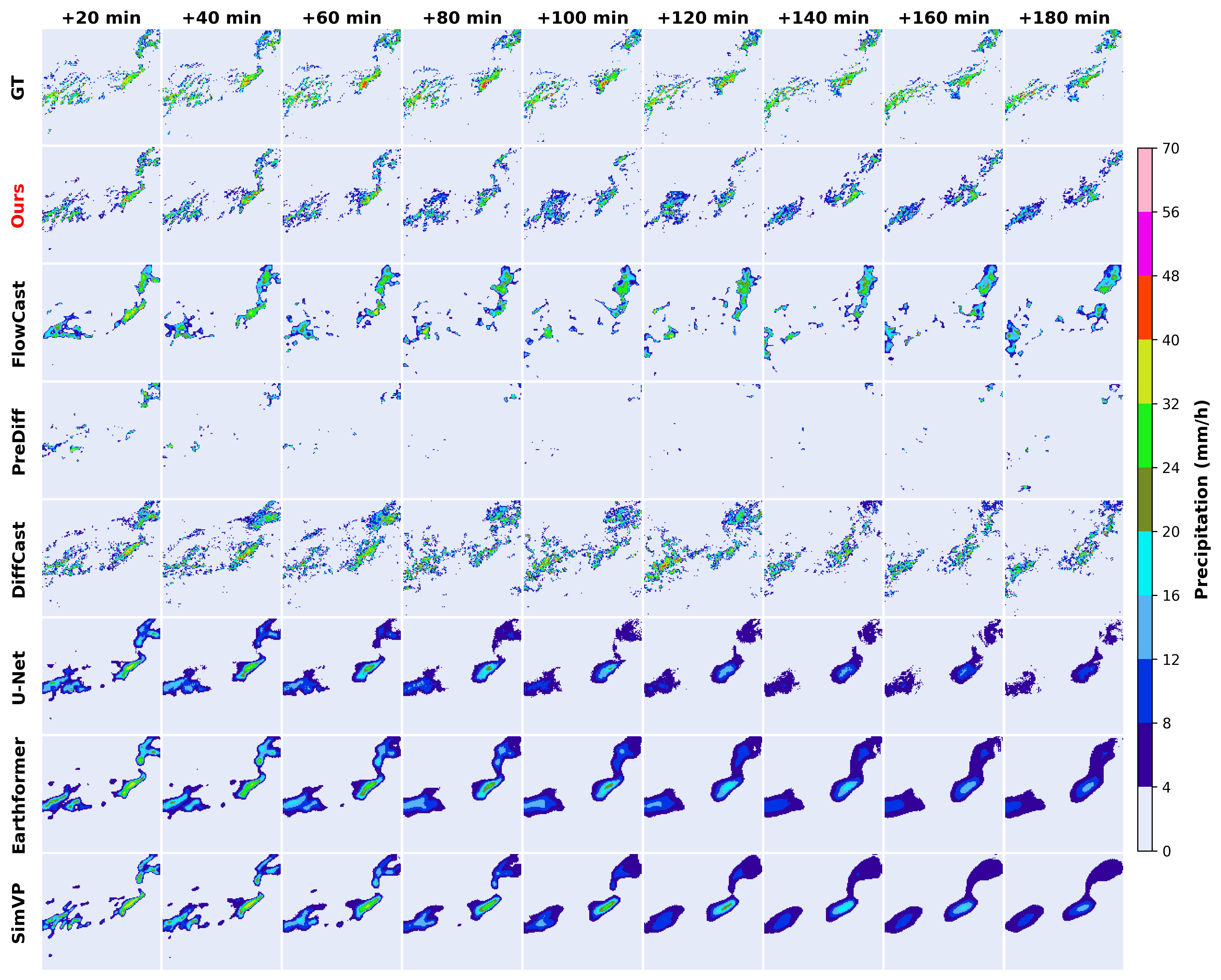}
        \caption{Additional qualitative comparison on the MeteoNet dataset (Example 5). Further evidence confirming that the proposed framework achieves superior generation fidelity, reliably tracking extreme precipitation peaks across complex and rapidly evolving meteorological sequences.}
        \label{fig:meteo_supp_example5}
    \end{minipage}
\end{figure*}

To further demonstrate the effectiveness and generalizability of PixelFlowCast across different geographical domains, 
we provide more visualization results on the SEVIR dataset 
in Figure~\ref{fig:supp_example1},
Figure~\ref{fig:supp_example2},
Figure~\ref{fig:supp_example3},
Figure~\ref{fig:supp_example4},
and Figure~\ref{fig:supp_example5},
and additional visual sequences on the MeteoNet dataset 
in Figure~\ref{fig:meteo_supp_example1},
Figure~\ref{fig:meteo_supp_example2},
Figure~\ref{fig:meteo_supp_example3},
Figure~\ref{fig:meteo_supp_example4},
and Figure~\ref{fig:meteo_supp_example5},
with samples displayed at intervals of 20 minutes.

The extended visual sequences clearly corroborate our quantitative results.
PixelFlowCast's direct pixel-space formulation intrinsically avoids latent-space smoothing to preserve fine radar textures.
This advantage is particularly evident at higher thresholds, where KANCondNet's enhanced nonlinear capacity
prevents detail loss and accurately resolves high-intensity precipitation cores.
Importantly, this robust performance persists over time;
across the 180-minute forecast horizon, the model maintains sharp boundaries and realistic distributions
without suffering from the progressive blurring characteristic of traditional approaches.

\clearpage

\end{document}